%% file: main.tex
\documentclass[runningheads]{llncs}
\usepackage{graphicx}
\usepackage{amsmath,amssymb} 
\usepackage{color}
\usepackage[width=122mm,left=12mm,paperwidth=146mm,height=193mm,top=12mm,paperheight=217mm]{geometry}

\usepackage[boxed]{algorithm2e}
\usepackage{caption}
\usepackage{mathtools}
\usepackage{amsfonts}
\usepackage{booktabs}
\usepackage{graphicx}
\usepackage{soul}
\usepackage{color}
\usepackage{xspace}
\usepackage[subrefformat=parens]{subfig}
\usepackage{soul}

\newlength{\soustable}
\makeatletter
\DeclareRobustCommand\onedot{\futurelet\@let@token\@onedot}
\def\@onedot{\ifx\@let@token.\else.\null\fi\xspace}
 
\def\ie{\emph{i.e}\onedot}

\def\etal{\emph{et al}\onedot}
\def\OURS{DeepCluster\xspace}

\makeatother

\begin{document}
\pagestyle{headings}
\mainmatter
\def\ECCV18SubNumber{1370}  

\title{Deep Clustering for Unsupervised Learning \\  of Visual Features} 

\titlerunning{Deep Clustering for Unsupervised Learning of Visual Features}

\authorrunning{Mathilde Caron \etal}

\author{Mathilde Caron \and Piotr Bojanowski \and Armand Joulin \and Matthijs Douze}
\institute{Facebook AI Research}

\maketitle

\begin{abstract}
  Clustering is a class of unsupervised learning methods that has been extensively applied and studied in computer vision.
  Little work has been done to adapt it to the end-to-end training of visual features on large scale datasets.
  In this work, we present \OURS, a clustering method that jointly learns the parameters of a neural network and the cluster assignments
  of the resulting features.
  \OURS iteratively groups the features with a standard clustering algorithm, $k$-means, and uses the subsequent assignments as supervision
  to update the weights of the network.
  We apply \OURS to the unsupervised training of convolutional neural networks on large datasets like ImageNet and YFCC100M.
  The resulting model outperforms the current state of the art by a significant margin on all the standard benchmarks.
\end{abstract}

\section{Introduction}

\input{intro.tex}

\section{Related Work}

\input{related.tex}

\section{Method}

\input{method.tex}

\section{Experiments}

\input{results.tex}

\input{discussion.tex}
\section{Conclusion}

In this paper, we propose a scalable clustering approach for the unsupervised learning of convnets.
It iterates between clustering  with $k$-means the features produced by the convnet and
updating its weights by predicting the cluster assignments as pseudo-labels in a discriminative loss.
If trained on large dataset like ImageNet or YFCC100M, it achieves performance that are significantly
better than the previous state-of-the-art on every standard transfer task.
Our approach makes little assumption about the inputs, and does not require much domain specific knowledge, making it a
good candidate to learn deep representations specific to domains where annotations are scarce.

\paragraph{Acknowledgement.}
We thank Alexandre Sablayrolles and the rest of the FAIR team for their feedback and fruitful discussion around this paper.
We would like to particularly thank Ishan Misra for spotting an error in our evaluation setting of Table~\ref{tab:linear}.

\clearpage

\bibliographystyle{splncs}
\bibliography{egbib}

\clearpage
\input{supp.tex}

\end{document}

%% file: intro.tex

Pre-trained convolutional neural networks, or convnets, have become the
building blocks in most computer vision
applications~\cite{ren2015faster,chen2016deeplab,weinzaepfel2013deepflow,carreira2016human}.
They produce excellent general-purpose features that can be used to improve the
generalization of models learned on a limited amount of
data~\cite{sharif2014cnn}.  The existence of ImageNet~\cite{deng2009imagenet},
a large fully-supervised dataset, has been fueling advances in pre-training of
convnets.  However, Stock and Cisse~\cite{stock2017convnets} have recently
presented empirical evidence that the performance of state-of-the-art classifiers on
ImageNet is largely underestimated, and little error is left unresolved.  This
explains in part why the performance has been saturating despite the numerous novel
architectures proposed in recent
years~\cite{chen2016deeplab,he2015delving,huang2016densely}.  As a matter of
fact, ImageNet is relatively small by today's standards; it ``only'' contains a
million images that cover the specific domain of object classification.  A
natural way to move forward is to build a bigger and more diverse dataset,
potentially consisting of billions of images.  This, in turn, would require a
tremendous amount of manual annotations, despite the expert knowledge in
crowdsourcing accumulated  by the community over the
years~\cite{kovashka2016crowdsourcing}.  Replacing labels by raw metadata
leads to biases in the visual representations with unpredictable
consequences~\cite{misra2016seeing}.  This calls for methods that can be
trained on internet-scale datasets with no supervision.

Unsupervised learning has been widely studied in the Machine Learning
community~\cite{friedman2001elements}, and algorithms for clustering,
dimensionality reduction or density estimation are regularly used in computer
vision
applications~\cite{turk1991face,shi2000normalized,joulin2010discriminative}.
For example, the ``bag of features'' model uses clustering on handcrafted local
descriptors to produce good image-level features~\cite{csurka2004visual}.  A
key reason for their success is that they can be applied on any specific domain
or dataset, like satellite or medical images, or on images captured with a new
modality, like depth, where annotations are not always available in quantity.
Several works have shown that it was possible to adapt unsupervised methods
based on density estimation or dimensionality reduction to deep
models~\cite{goodfellow2014generative,kingma2013auto}, leading to promising
all-purpose visual
features~\cite{bojanowski2017unsupervised,donahue2016adversarial}.  Despite the
primeval success of clustering approaches in image classification, very few
works~\cite{yang2016joint,xie2016unsupervised} have been proposed to adapt them
to the end-to-end training of convnets, and never at scale.  An issue is that
clustering methods have been primarily designed for linear models on top of
fixed features, and they scarcely work if the features have to be learned
simultaneously.  For example, learning a convnet with $k$-means would lead to a
trivial solution where the features are zeroed, and the clusters are collapsed
into a single entity.

\begin{figure}[!t]
  \centering
  \includegraphics[width=.9\linewidth]{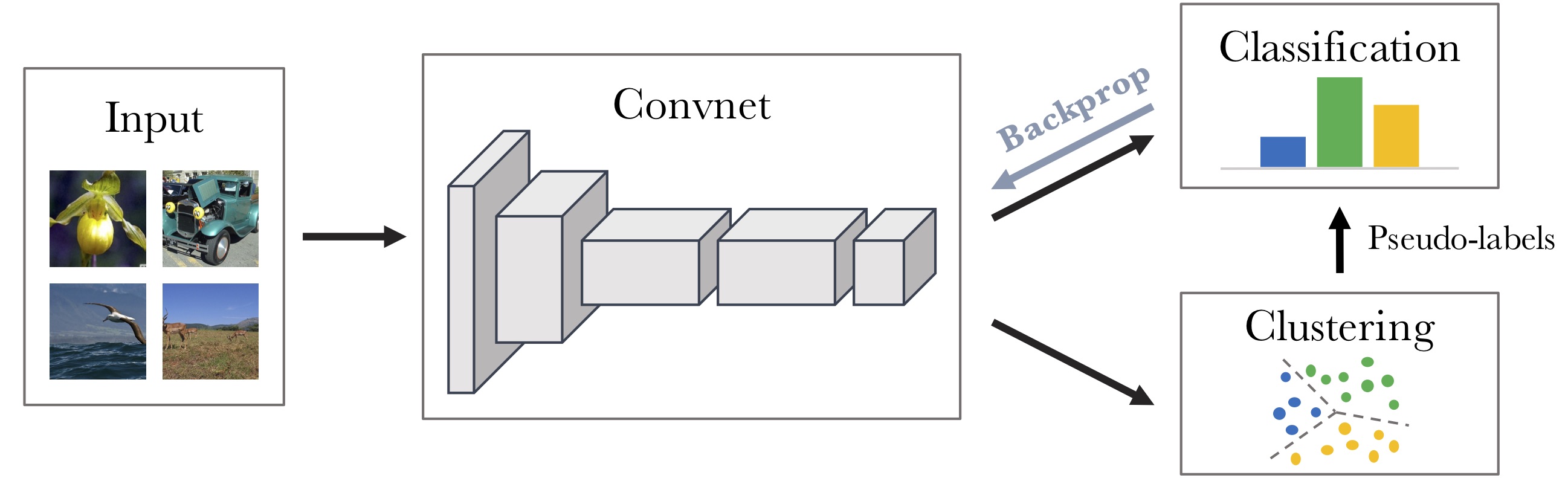}
  \caption{
    Illustration of the proposed method: we iteratively cluster deep features and use the cluster assignments as pseudo-labels to learn the parameters of the convnet.
  }
  \label{fig:pullfig}
\end{figure}

In this work, we propose a novel clustering approach for the large scale
end-to-end training of convnets.  We show that it is possible to obtain useful
general-purpose visual features with a clustering framework.  Our approach,
summarized in Figure~\ref{fig:pullfig}, consists in alternating between
clustering of the image descriptors and updating the weights of the convnet by
predicting the cluster assignments.  For simplicity, we focus our study on
$k$-means, but other clustering approaches can be used, like Power Iteration
Clustering~(PIC)~\cite{lin2010power}.  The overall pipeline is sufficiently
close to the standard supervised training of a convnet to reuse many common
tricks~\cite{ioffe2015batch}.  Unlike self-supervised
methods~\cite{doersch2015unsupervised,noroozi2016unsupervised,pathak2016learning},
clustering has the advantage of requiring little domain knowledge and no specific
signal from the
inputs~\cite{zhang2016colorful,wang2015unsupervised}.  Despite its simplicity,
our approach achieves significantly higher performance than previously
published unsupervised methods on both ImageNet classification and transfer tasks.

Finally, we probe the robustness of our framework by modifying the experimental protocol,
in particular the training set and the convnet architecture. The resulting set of experiments extends the
discussion initiated by Doersch~\etal~\cite{doersch2015unsupervised} on the impact of these choices on
the performance of unsupervised methods.
We demonstrate that our
approach is robust to a change of architecture.
Replacing an AlexNet by a
VGG~\cite{simonyan2014very} significantly improves the quality of the features
and their subsequent transfer performance.  More importantly, we discuss
the use of ImageNet as a training set for unsupervised models.  While it helps
understanding the impact of the labels on the performance of a network,
ImageNet has a particular image distribution inherited from its use for a
fine-grained image classification challenge: it is composed of well-balanced classes and contains
a wide variety of dog breeds for example.  We consider, as an alternative, random Flickr
images from the YFCC100M dataset of Thomee~\etal~\cite{thomee2015new}.  We show
that our approach maintains state-of-the-art performance when trained on this
uncured data distribution.  Finally,  current benchmarks focus on
the capability of unsupervised convnets to capture class-level
information.  We propose to also evaluate them on image retrieval
benchmarks to measure their capability to capture instance-level
information.

In this paper, we make the following contributions:
\textbf{(i)} a novel unsupervised method for the end-to-end learning of convnets that works with any standard clustering algorithm, like $k$-means, and requires minimal additional steps;
\textbf{(ii)} state-of-the-art performance on many standard transfer tasks used in unsupervised learning;
\textbf{(iii)} performance above the previous state of the art when trained on an uncured image distribution;
\textbf{(iv)} a discussion about the current evaluation protocol in unsupervised feature learning.

%% file: related.tex


\noindent\textbf{Unsupervised learning of features.}
Several approaches related to our work learn deep models with no supervision.
Coates and Ng~\cite{coates2012learning} also use $k$-means to pre-train convnets, but learn each layer sequentially
in a bottom-up fashion, while we do it in an end-to-end fashion.
Other clustering losses~\cite{yang2016joint,xie2016unsupervised,dosovitskiy2014discriminative,liao2016learning} 
have been considered to jointly learn 
convnet features and image clusters but they have never been tested on a scale to allow a thorough study on modern convnet architectures.
Of particular interest,
Yang~\etal~\cite{yang2016joint} iteratively learn convnet features and clusters with a recurrent framework.
Their model offers promising performance on small datasets but may be 
challenging to scale to the number of images required for convnets to be competitive.
Closer to our work, Bojanowski and Joulin~\cite{bojanowski2017unsupervised} learn visual features on a large dataset with
a loss that attempts to preserve the information flowing through the network~\cite{linsker1988towards}. 
Their approach discriminates between images in a similar way as examplar SVM~\cite{malisiewicz2011ensemble}, while we are simply clustering them.
\\

\noindent\textbf{Self-supervised learning.}
A popular form of unsupervised learning, called ``self-supervised learning''~\cite{de1994learning},
uses pretext tasks to replace the labels annotated by humans by ``pseudo-labels'' directly computed from the raw input data.
For example,
Doersch~\etal~\cite{doersch2015unsupervised} use the prediction of the relative position of patches in an image as a pretext task,
while Noroozi and Favaro~\cite{noroozi2016unsupervised} train a network to spatially rearrange shuffled patches.
Another use of spatial cues is the work of Pathak~\etal~\cite{pathak2016context} where missing pixels are guessed based
on their surrounding.
Paulin~\etal~\cite{paulin2015local} learn patch level Convolutional Kernel Network~\cite{mairal2014convolutional} using an image retrieval setting. 
Others leverage the temporal signal available in videos by 
predicting the camera transformation between consecutive frames~\cite{agrawal2015learning}, exploiting the temporal coherence of tracked patches~\cite{wang2015unsupervised} or segmenting video based on motion~\cite{pathak2016learning}. 
Appart from spatial and temporal coherence, many other signals have been explored: image colorization~\cite{zhang2016colorful,larsson2016learning}, cross-channel prediction~\cite{zhang2016split}, sound~\cite{owens2016ambient} or instance counting~\cite{noroozi2017representation}.
More recently, several strategies for combining multiple cues have been proposed~\cite{wang2017transitive,doersch2017multi}.
Contrary to our work, these approaches are domain dependent, requiring expert knowledge 
to carefully design a pretext task that may lead to transferable features.
\\

\noindent\textbf{Generative models.}
Recently, unsupervised learning has been making a lot of progress on image generation.
Typically, a parametrized mapping is learned between a predefined random noise and the images,
with either an autoencoder~\cite{kingma2013auto,bengio2007greedy,huang2007unsupervised,masci2011stacked,vincent2010stacked}, 
a generative adversarial network (GAN)~\cite{goodfellow2014generative} 
or more directly with a reconstruction loss~\cite{bojanowski2017optimizing}.
Of particular interest, 
the discriminator of a GAN can produce visual features, but their performance are relatively disappointing~\cite{donahue2016adversarial}. 
Donahue~\etal~\cite{donahue2016adversarial} and Dumoulin~\etal~\cite{dumoulin2016adversarially} have shown that adding an encoder
to a GAN produces visual features that are much more competitive.

%% file: method.tex
After a short introduction to the supervised learning of convnets, we describe our unsupervised approach as well as the specificities of its optimization.

\subsection{Preliminaries}
Modern approaches to computer vision, based on statistical learning, require good image featurization.
In this context, convnets are a popular choice for mapping raw images to a vector space of fixed dimensionality.
When trained on enough data, they constantly achieve the best performance on standard classification benchmarks~\cite{he2015delving,krizhevsky2012imagenet}.
We denote by $f_\theta$ the convnet mapping, where $\theta$ is the set of corresponding parameters.
We refer to the vector obtained by applying this mapping to an image as feature or representation.
Given a training set $X = \{x_1, x_2, \dots , x_N\}$ of $N$ images, we want to find a parameter $\theta^*$ such that the mapping $f_{\theta^*}$ produces good general-purpose features.

These parameters are traditionally learned with supervision, \ie each image $x_n$ is associated with a label $y_n$ in $\{0,1\}^k$.
This label represents the image's membership to one of $k$ possible predefined classes.
A parametrized classifier $g_W$ predicts the correct labels on top of the features $f_\theta(x_n)$.
The parameters $W$ of the classifier and the parameter $\theta$ of the mapping are then jointly learned by optimizing the following problem:
\begin{eqnarray}\label{eq:sup}
  \min_{\theta, W} \frac{1}{N} \sum_{n=1}^N \ell\left(g_W\left( f_\theta(x_n) \right), y_n\right),
\end{eqnarray}
where $\ell$ is the multinomial logistic loss, also known as the negative log-softmax function.
This cost function is minimized using mini-batch stochastic gradient descent~\cite{bottou2012stochastic} and backpropagation to compute the gradient~\cite{lecun1998gradient}.

\subsection{Unsupervised learning by clustering}
When $\theta$ is sampled from a Gaussian distribution, without any learning, $f_\theta$ does not produce good features.
However the performance of such random features on standard transfer tasks, is far above the chance level.
For example, a multilayer perceptron classifier on top of the last convolutional layer of a random AlexNet achieves 12\% in accuracy on ImageNet while the chance is at $0.1\%$~\cite{noroozi2016unsupervised}.
The good performance of random convnets is intimately tied to their convolutional structure which gives a strong prior on the input signal.
The idea of this work is to exploit this weak signal to bootstrap the discriminative power of a convnet.
We cluster the output of the convnet and use the subsequent cluster assignments as ``pseudo-labels'' to optimize Eq.~(\ref{eq:sup}).
This deep clustering (DeepCluster) approach iteratively learns the features and groups them.

Clustering has been widely studied and many approaches have been developed for a variety of circumstances. In the absence of points of comparisons,
we focus on a standard clustering algorithm, $k$-means. Preliminary results with other clustering algorithms indicates that this choice is not crucial.
$k$-means takes a set of vectors as input, in our case the features $f_\theta(x_n)$ produced by the convnet, and clusters them into $k$ distinct groups based on a geometric criterion.
More precisely, it jointly learns a $d\times k$ centroid matrix $C$ and the cluster assignments $y_n$ of each image $n$ by solving the following problem:
\begin{equation}
\label{eq:kmeans}
  \min_{C \in \mathbb{R}^{d\times k}}
  \frac{1}{N}
  \sum_{n=1}^N
  \min_{y_n \in \{0,1\}^{k}}
  \| f_\theta(x_n) -  C y_n \|_2^2
  \quad
  \text{such that}
  \quad
  y_n^\top 1_k = 1.
\end{equation}
Solving this problem provides a set of optimal assignments $(y_n^*)_{n\le N}$ and a centroid matrix $C^*$.
These assignments are then used as pseudo-labels; we make no use of the centroid matrix.

Overall, \OURS alternates between clustering the features to produce pseudo-labels using Eq.~(\ref{eq:kmeans})
and updating the parameters of the convnet by predicting these pseudo-labels using Eq.~(\ref{eq:sup}).
This type of alternating procedure is prone to trivial solutions; we describe how to avoid such degenerate solutions in the next section.

\subsection{Avoiding trivial solutions}
The existence of trivial solutions is not specific to the unsupervised training of neural networks, but to any method that jointly learns a discriminative classifier and the labels.
Discriminative clustering suffers from this issue even when applied to linear models~\cite{xu2005maximum}.
Solutions are typically based on constraining or penalizing the minimal number of points per cluster~\cite{bach2008diffrac,joulin2012convex}.
These terms are computed over the whole dataset, which is not applicable to the training of convnets on large scale datasets.
In this section, we briefly describe the causes of these trivial solutions and give simple and scalable workarounds.
\\

\noindent\textbf{Empty clusters.}
A discriminative model learns decision boundaries between classes.
An optimal decision boundary is to assign all of the inputs to a single cluster~\cite{xu2005maximum}.
This issue is caused by the absence of mechanisms to prevent from empty clusters and arises in linear models as much as in convnets.
A common trick used in feature quantization~\cite{johnson2017billion} consists in automatically reassigning empty clusters during the $k$-means optimization.
More precisely, when a cluster becomes empty, we randomly select a non-empty cluster and use its centroid with a small random perturbation as the new centroid for the empty cluster.
We then reassign the points belonging to the non-empty cluster to the two resulting clusters.
\\

\noindent\textbf{Trivial parametrization.}
If the vast majority of images is assigned to a few clusters, the parameters $\theta$ will exclusively discriminate between them.
In the most dramatic scenario where all but one cluster are singleton, minimizing Eq.~(\ref{eq:sup}) leads to a
trivial parametrization where the convnet will predict the same output regardless of the input.
This issue also arises in supervised classification when the number of images per class is highly unbalanced.
For example, metadata, like hashtags, exhibits a Zipf distribution, with a few labels dominating the whole distribution~\cite{joulin2016learning}.
A strategy to circumvent this issue is to sample images based on a uniform distribution over the classes, or pseudo-labels.
This is equivalent to weight the contribution of an input to the loss function in Eq.~(\ref{eq:sup}) by the inverse of the size of its assigned cluster.

\subsection{Implementation details}

\noindent\textbf{Convnet architectures.}
For comparison with previous works, we use a standard AlexNet~\cite{krizhevsky2012imagenet} architecture.
It consists of five convolutional layers with $96$, $256$, $384$, $384$ and $256$ filters; and of three fully connected layers.
We remove the Local Response Normalization layers and use batch normalization~\cite{ioffe2015batch}.
We also consider a VGG-16~\cite{simonyan2014very} architecture with batch normalization.
Unsupervised methods often do not work directly on color and different strategies have been considered as alternatives
~\cite{doersch2015unsupervised,noroozi2016unsupervised}.
We apply a fixed linear transformation based on Sobel filters to remove color and increase local contrast~\cite{bojanowski2017unsupervised,paulin2015local}.
\\

\noindent\textbf{Training data.}
We train \OURS on ImageNet~\cite{deng2009imagenet} unless mentioned otherwise.
It contains~$1.3$M images uniformly distributed into~$1,000$ classes.
\\

\noindent\textbf{Optimization.}
We cluster the central cropped images features and perform data augmentation (random horizontal flips and crops of random sizes and aspect ratios) when training the network. 
This enforces invariance to data augmentation which is useful for feature learning~\cite{dosovitskiy2014discriminative}.
The network is trained with dropout~\cite{srivastava2014dropout}, a constant step size, an~$\ell_2$
penalization of the weights~$\theta$ and a momentum of~$0.9$.
Each mini-batch contains $256$ images.
For the clustering, features are PCA-reduced to $256$ dimensions, whitened and $\ell_2$-normalized.
We use the $k$-means implementation of Johnson~\etal~\cite{johnson2017billion}.
Note that running k-means takes a third of the time because a forward pass on the full dataset is needed.
One could reassign the clusters every $n$ epochs, but we found out that our setup on ImageNet (updating the clustering every epoch) was nearly optimal.
On Flickr, the concept of epoch disappears: choosing the tradeoff between the parameter updates and the cluster reassignments is more subtle.
We thus kept almost the same setup as in ImageNet.
We train the models for $500$ epochs, which takes $12$ days on a Pascal P100 GPU for AlexNet.
\\

\noindent\textbf{Hyperparameter selection.}
We select hyperparameters on a down-stream task,~\ie, object classification on
the validation set of \textsc{Pascal} VOC with no fine-tuning.
We use the publicly available code of Kr\"ahenb\"uhl\footnote{https://github.com/philkr/voc-classification}.

%% file: results.tex

In a preliminary set of experiments, we study the behavior of \OURS during training.
We then qualitatively assess the filters learned with \OURS before comparing our approach to previous state-of-the-art models on standard benchmarks.

\begin{figure}[t]
	\centering
	\subfloat[Clustering quality]{
    \includegraphics[width=0.32\linewidth]{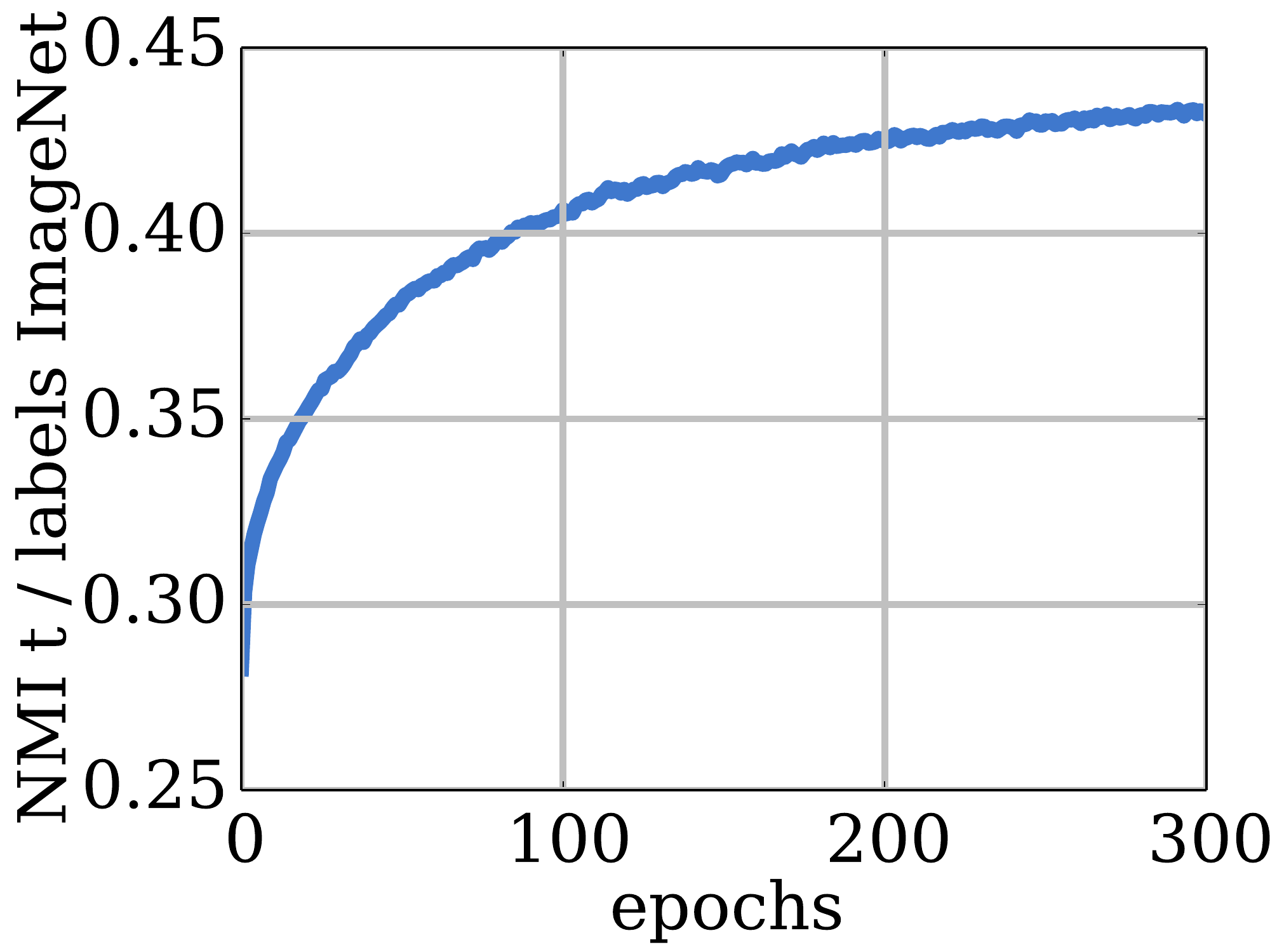}
		\label{fig:edges}
	}
	\subfloat[Cluster reassignment]{
    \includegraphics[width=0.32\linewidth]{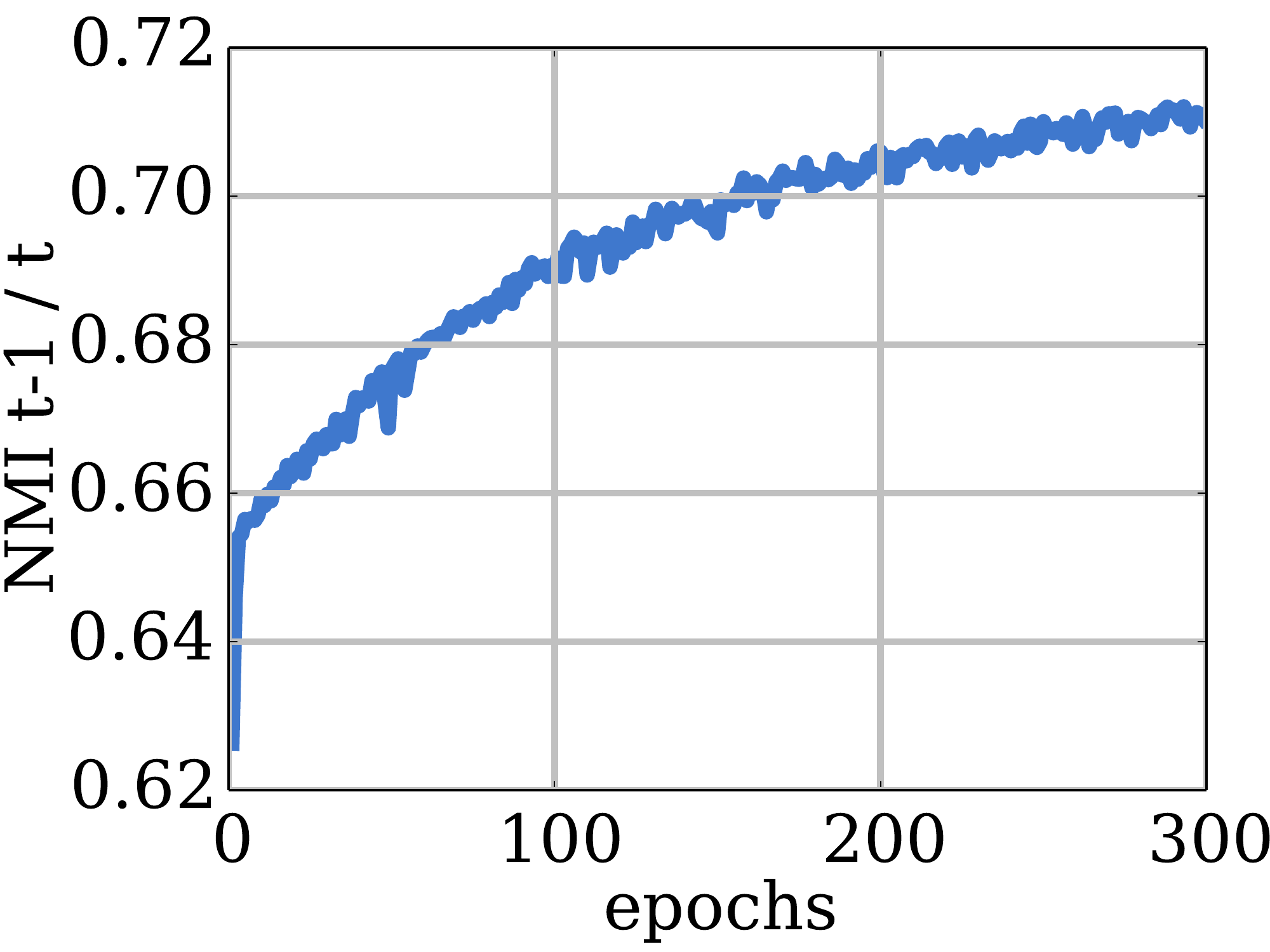}
		\label{fig:reassing}
	}
	\subfloat[Influence of k]{
    \includegraphics[width=0.32\linewidth]{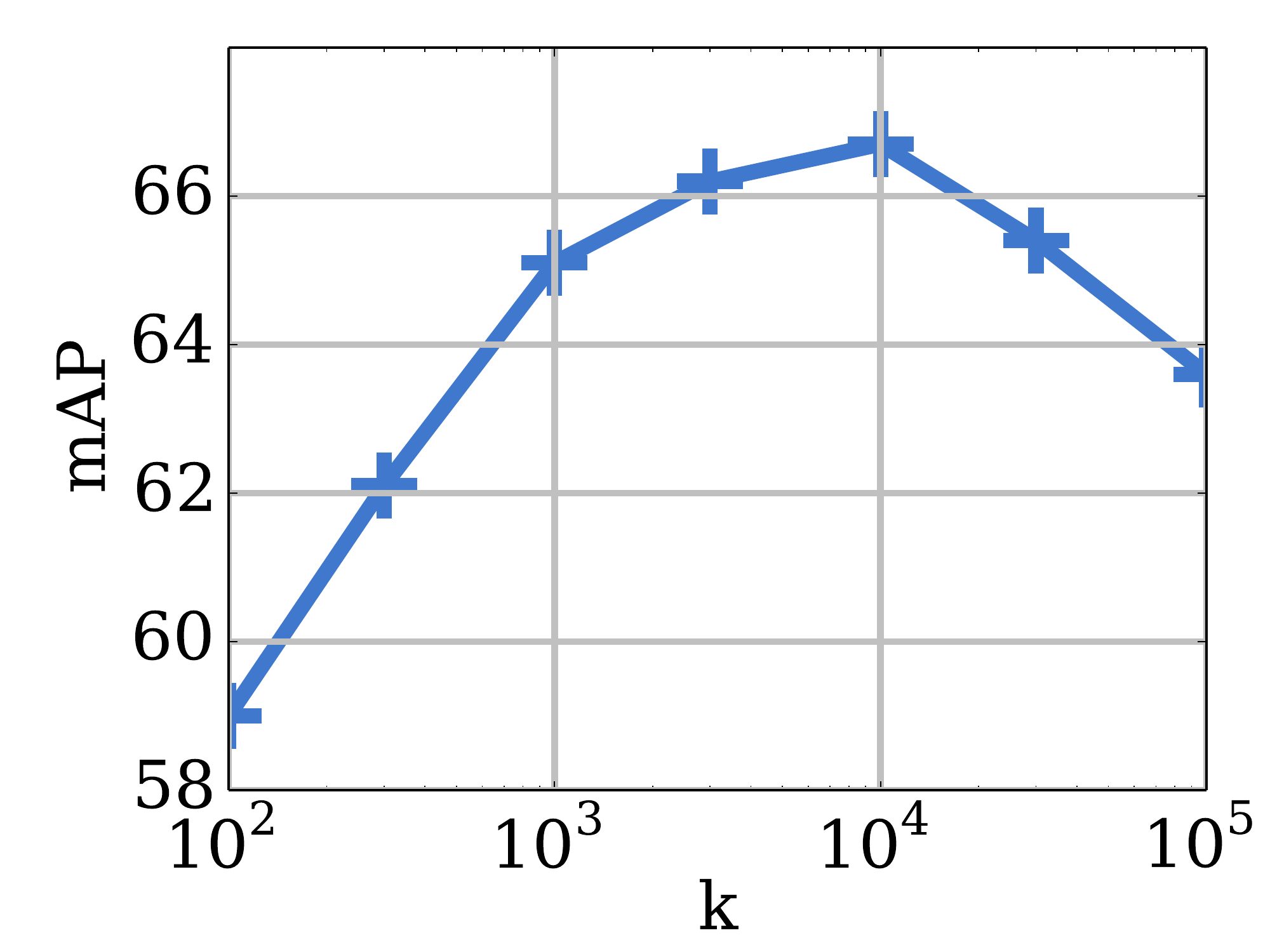}
		\label{fig:choice-k}
	}
	\caption{
		Preliminary studies.
		\protect\subref{fig:edges}: evolution of the clustering quality along training epochs;
		\protect\subref{fig:reassing}: evolution of cluster reassignments at each clustering step;
		\protect\subref{fig:choice-k}: validation mAP classification performance for various choices of $k$.
	}
	\label{fig:prelim}
\end{figure}


\begin{figure}[t]
  \centering
  \begin{tabular}{cc}
    \includegraphics[width=0.48\linewidth]{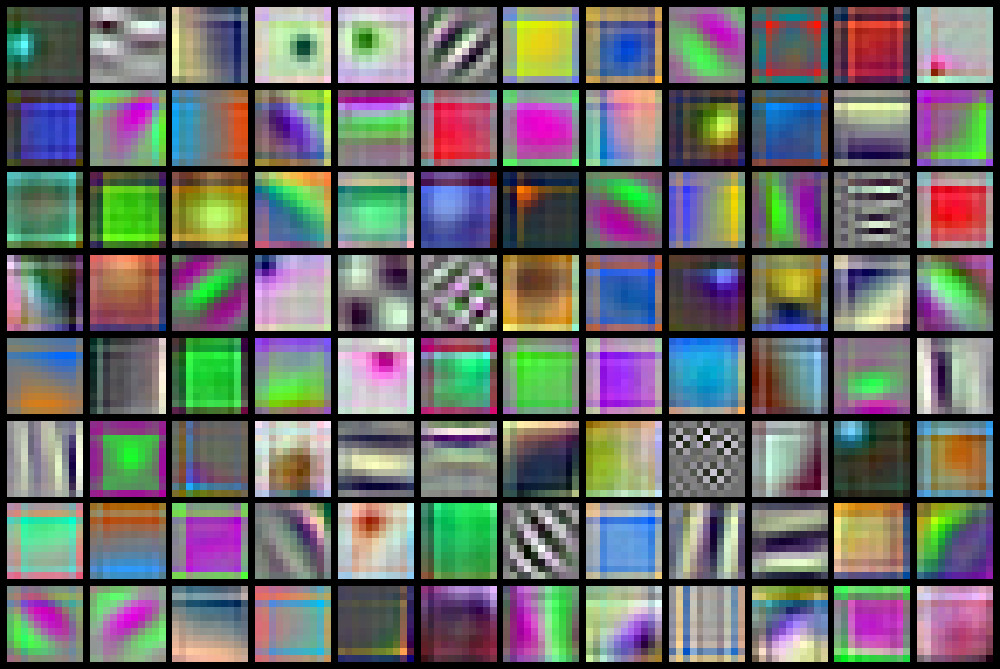}&
    \includegraphics[width=0.48\linewidth]{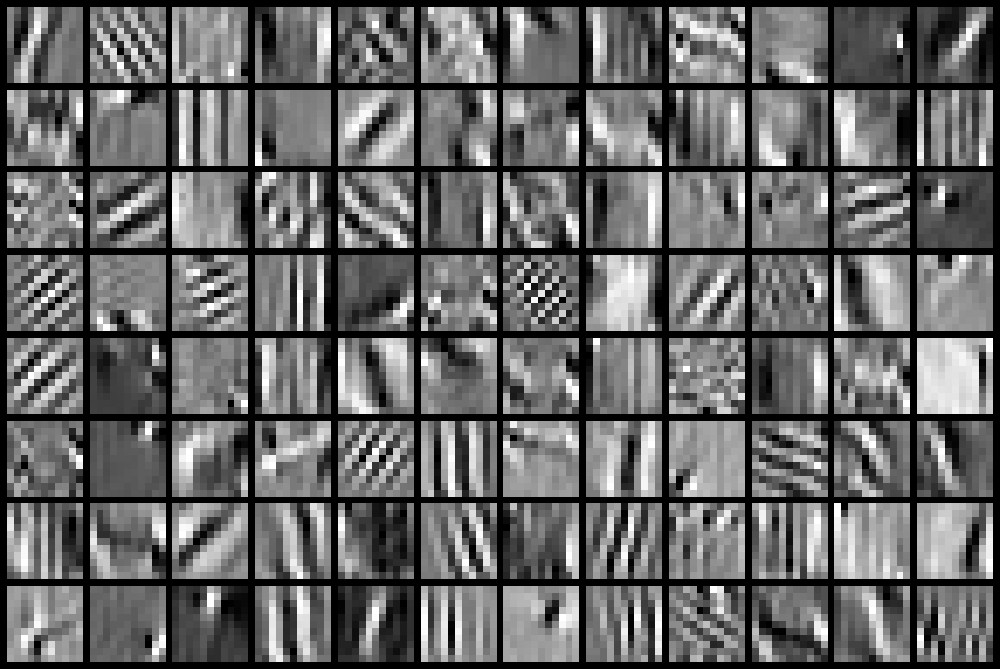}
  \end{tabular}
  \caption{Filters from the first layer of an AlexNet trained on unsupervised ImageNet on raw RGB input (left) or after a Sobel filtering (right).
  }
  \label{fig:filters}
\end{figure}

\subsection{Preliminary study}

We measure the information shared between two different assignments $A$ and $B$ of the same data by the
Normalized Mutual Information (NMI), defined as:
\begin{equation*}
\mathrm{NMI}(A;B)=\frac{\mathrm{I}(A;B)}{\sqrt{\mathrm{H}(A) \mathrm{H}(B)}}
\end{equation*}
where $\mathrm{I}$ denotes the mutual information and $\mathrm{H}$ the entropy. This measure can be applied to any assignment coming
from the clusters or the true labels.
If the two assignments $A$ and $B$ are independent, the NMI is equal to $0$.
If one of them is deterministically predictable from the other, the NMI is equal to $1$.
\\

\noindent\textbf{Relation between clusters and labels.}
Figure~\subref*{fig:edges} shows the evolution of the NMI between the cluster assignments and the ImageNet labels during training.
It measures the capability of the model to predict class level information.
Note that we only use this measure for this analysis and not in any model selection process.
The dependence between the clusters and the labels increases over time, showing that our features progressively capture information
related to object classes.
\\

\noindent\textbf{Number of reassignments between epochs.}
At each epoch, we reassign the images to a new set of clusters, with no guarantee of stability.
Measuring the NMI between the clusters at epoch $t-1$ and $t$ gives an insight on the actual
stability of our model.
Figure~\subref*{fig:reassing} shows the evolution of this measure during training.
The NMI is increasing, meaning that there are less and less reassignments and the clusters are stabilizing over time.
However, NMI saturates below $0.8$, meaning that a significant fraction of images are regularly reassigned between epochs.
In practice, this has no impact on the training and the models do not diverge.
\\

\noindent\textbf{Choosing the number of clusters.}
We measure the impact of the number $k$ of clusters used in $k$-means on the quality of the model.
We report the same down-stream task as in the hyperparameter selection process, \ie mAP on the \textsc{Pascal} VOC 2007 classification validation set.
We vary $k$ on a logarithmic scale, and report results after $300$ epochs in Figure~\subref*{fig:choice-k}.
The performance after the same number of epochs for every $k$ may not be directly comparable, but it reflects the hyper-parameter selection process
used in this work.
The best performance is obtained with $k=10,000$.
Given that we train our model on ImageNet, one would expect $k = 1000$ to yield the best results, but apparently some amount of over-segmentation is beneficial.


\subsection{Visualizations}
\label{sec:viz}

\noindent\textbf{First layer filters.}
Figure~\ref{fig:filters} shows the filters from the first layer of an AlexNet trained with \OURS on raw RGB images and images preprocessed with a Sobel filtering.
The difficulty of learning convnets on raw images has been noted before~\cite{bojanowski2017unsupervised,doersch2015unsupervised,noroozi2016unsupervised,paulin2015local}.
As shown in the left panel of Fig.~\ref{fig:filters}, most filters capture only color information that typically plays a little role for object classification~\cite{van2011evaluating}.
Filters obtained with Sobel preprocessing act like edge detectors.
\\

\begin{figure}[t]
\centering
\begin{tabular}{cccccccc}
  \multicolumn{2}{c}{\texttt{conv1}} &~~~& \multicolumn{2}{c}{\texttt{conv3}} &~~~& \multicolumn{2}{c}{\texttt{conv5}}\\
  \includegraphics[width=0.15\linewidth]{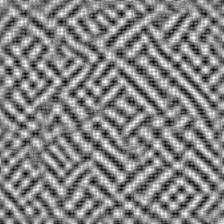}&
  \includegraphics[width=0.15\linewidth]{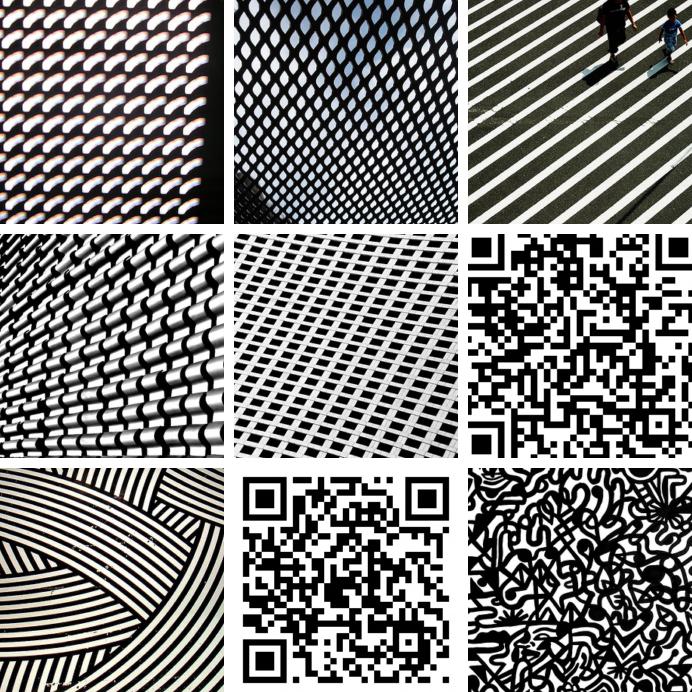}&&
  \includegraphics[width=0.15\linewidth]{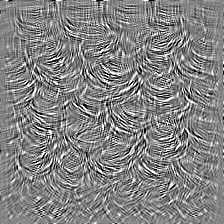}&
  \includegraphics[width=0.15\linewidth]{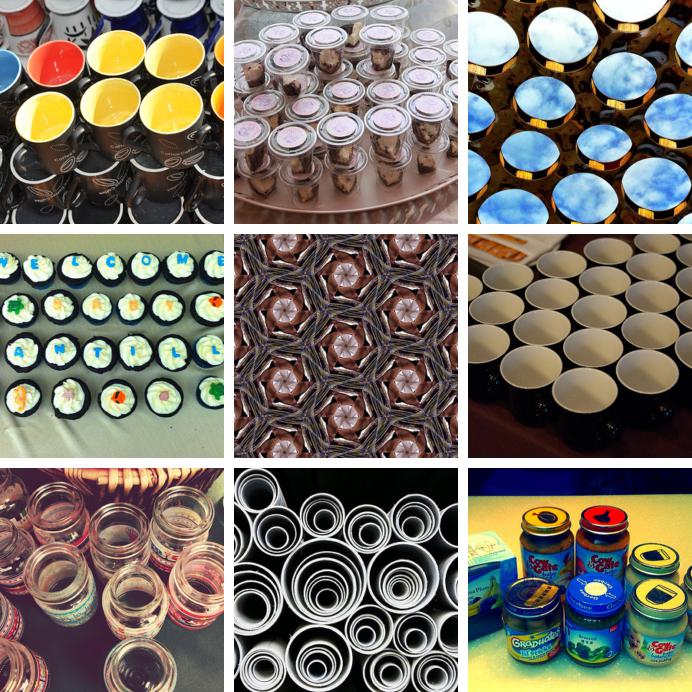}&&
  \includegraphics[width=0.15\linewidth]{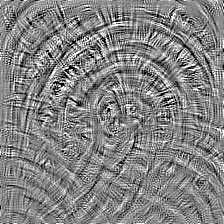}&
  \includegraphics[width=0.15\linewidth]{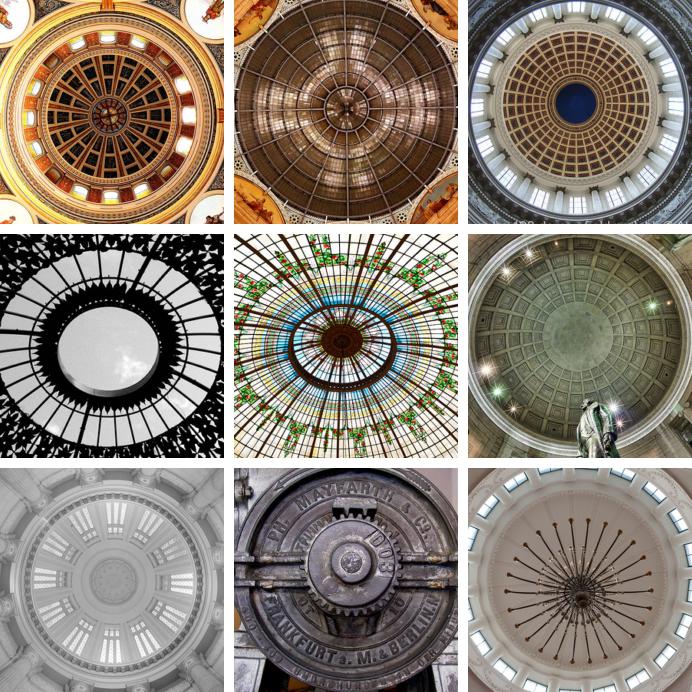}
\\
  \includegraphics[width=0.15\linewidth]{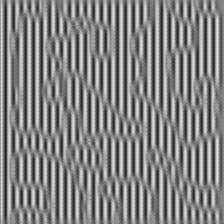}&
  \includegraphics[width=0.15\linewidth]{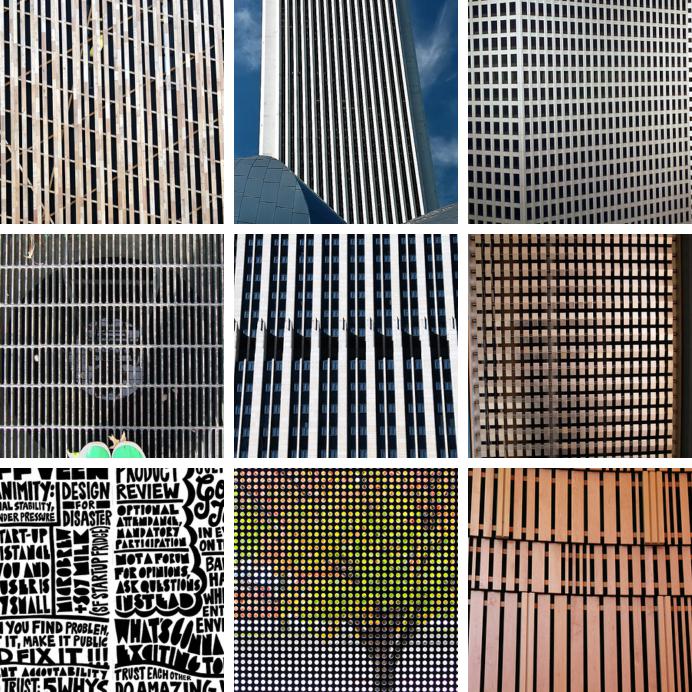}&&
  \includegraphics[width=0.15\linewidth]{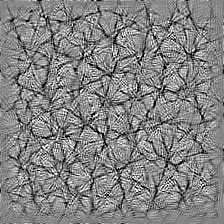}&
  \includegraphics[width=0.15\linewidth]{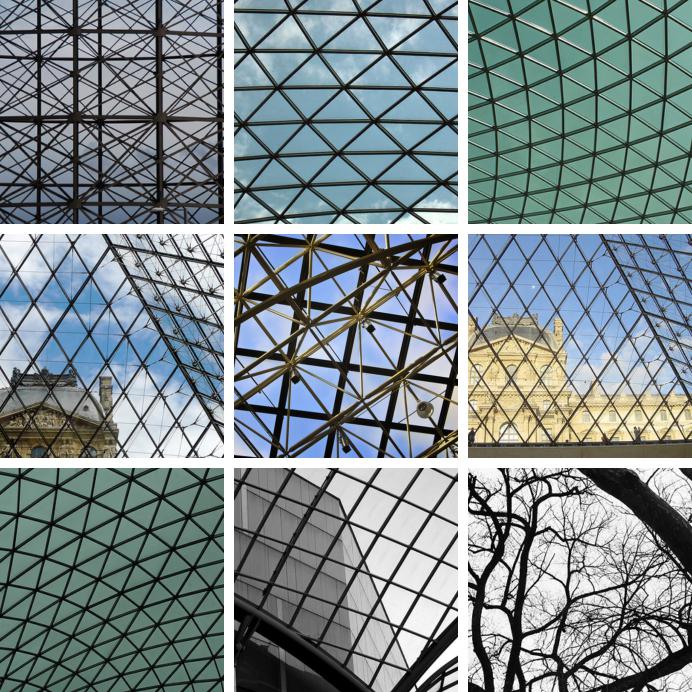}&&
  \includegraphics[width=0.15\linewidth]{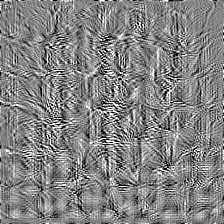}&
  \includegraphics[width=0.15\linewidth]{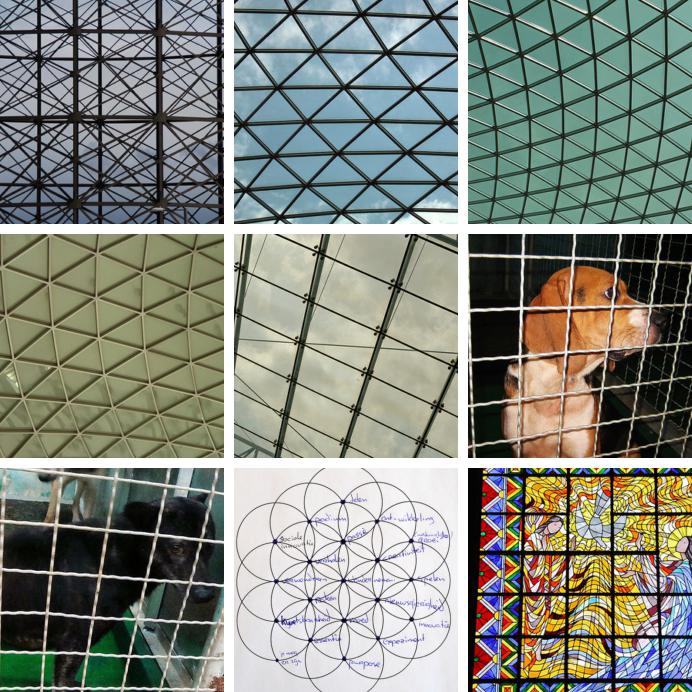}
\end{tabular}
\caption{
  Filter visualization and top $9$ activated images from a subset of $1$ million images from YFCC100M for target filters in the
  layers \texttt{conv1}, \texttt{conv3} and \texttt{conv5} of an AlexNet trained with \OURS on ImageNet.
  The filter visualization is obtained by learning an input image that maximizes the response to a target filter~\cite{yosinski2015understanding}.
}
\label{fig:activ}
\end{figure}

\begin{figure}[t]
\centering
\begin{tabular}{ccccccc}
Filter $0$ && Filter $33$ && Filter $145$ && Filter $194$
\\
\includegraphics[width=0.24\linewidth]{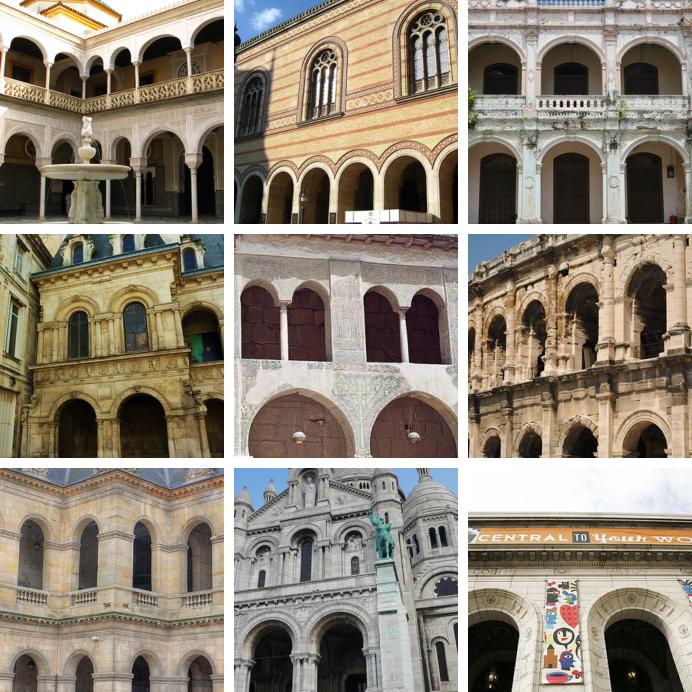}&&
\includegraphics[width=0.24\linewidth]{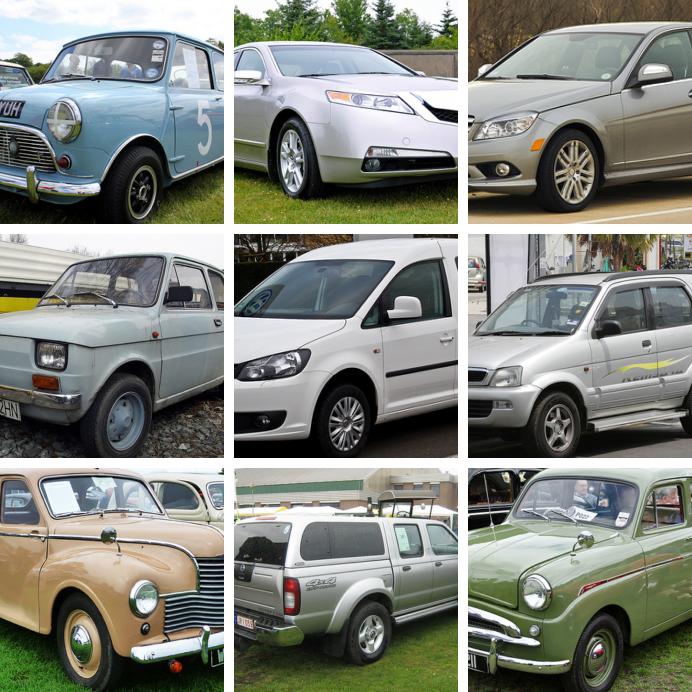}&&
\includegraphics[width=0.24\linewidth]{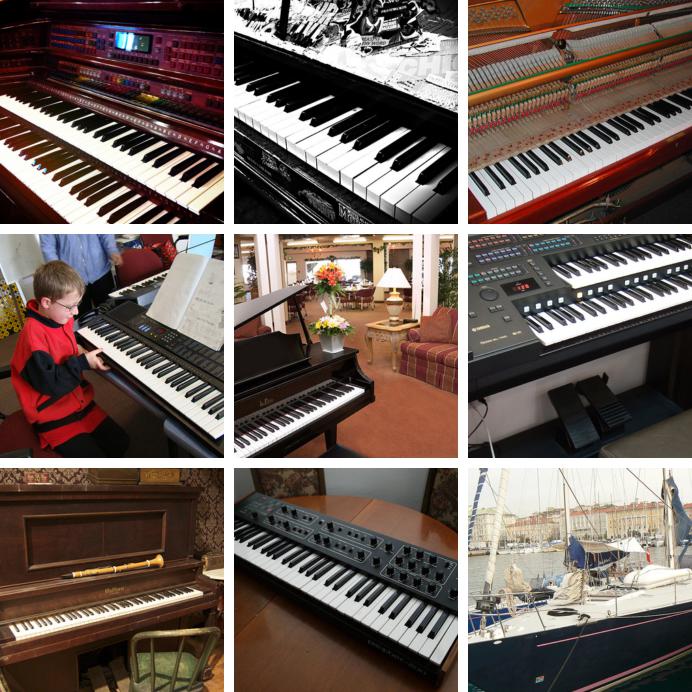}&&
\includegraphics[width=0.24\linewidth]{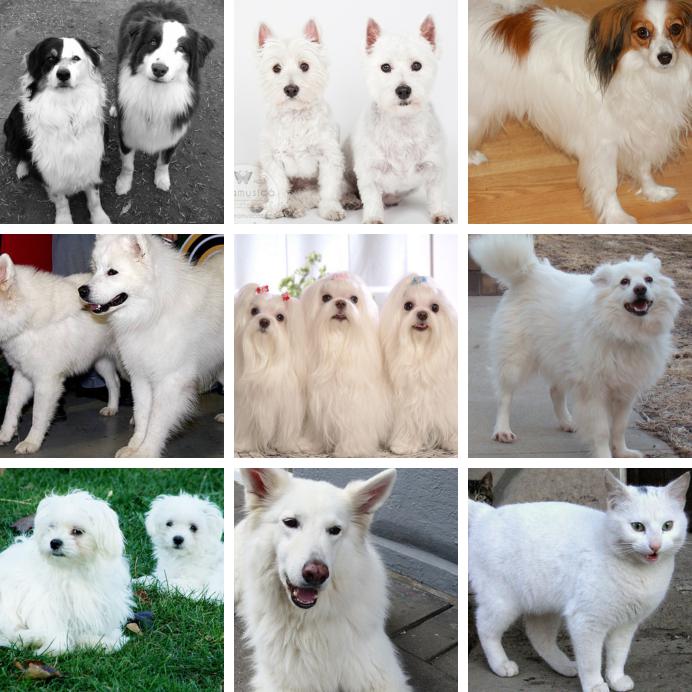}
\\
Filter $97$ && Filter $116$ && Filter $119$ && Filter $182$
\\
\includegraphics[width=0.24\linewidth]{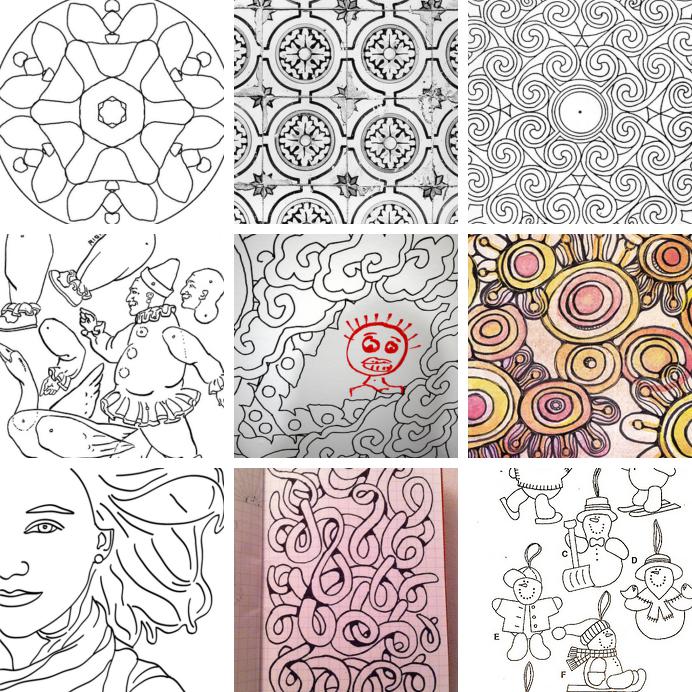}&&
\includegraphics[width=0.24\linewidth]{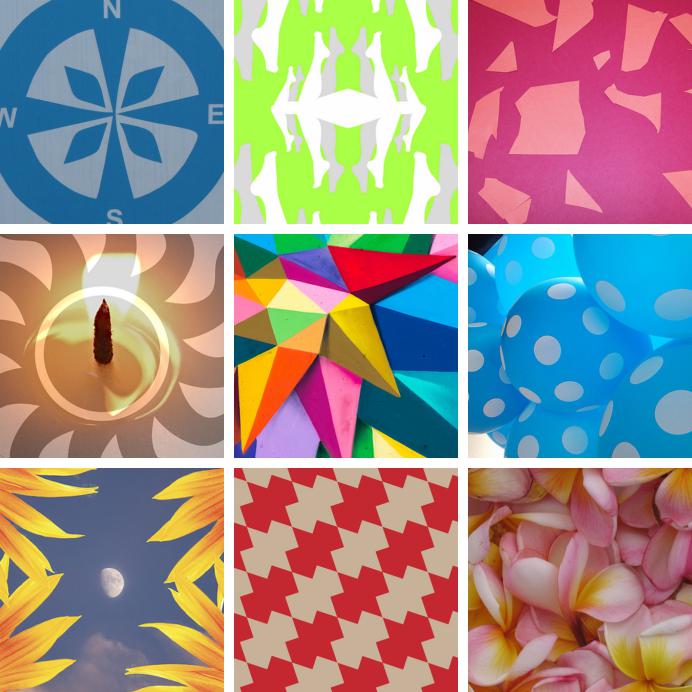}&&
\includegraphics[width=0.24\linewidth]{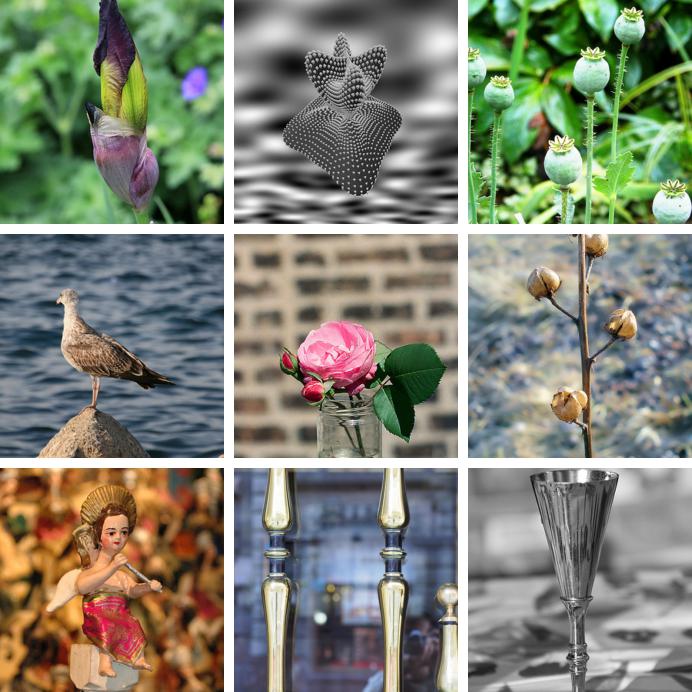}&&
\includegraphics[width=0.24\linewidth]{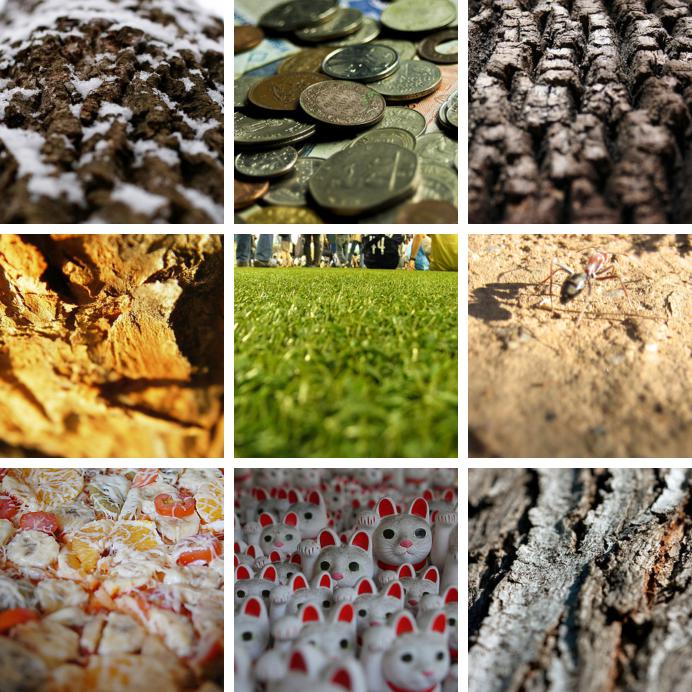}
\end{tabular}
\caption{
  Top $9$ activated images from a random subset of $10$ millions images from YFCC100M for target filters in the last convolutional layer.
  The top row corresponds to filters sensitive to activations by images containing objects.
  The bottom row exhibits filters more sensitive to stylistic effects.
  For instance, the filters $119$ and $182$ seem to be respectively excited by background blur and depth of field effects.
}
\label{fig:waouh}
\end{figure}

\noindent\textbf{Probing deeper layers.}
We assess the quality of a target filter by learning an input image that maximizes its activation~\cite{erhan2009visualizing,zeiler2014visualizing}.
We follow the process described by Yosinki~\etal~\cite{yosinski2015understanding} with a
cross entropy function between the target filter and the other filters of the same layer.
Figure~\ref{fig:activ} shows these synthetic images as well as the $9$ top activated images from a subset of $1$ million images
from YFCC100M.
As expected, deeper layers in the network seem to capture larger textural structures.
However, some filters in the last convolutional layers seem to be simply replicating
the texture already captured in previous layers, as shown on the second row of Fig.~\ref{fig:waouh}.
This result corroborates the observation by Zhang~\etal~\cite{zhang2016split} that features
from \texttt{conv3} or \texttt{conv4} are more discriminative than those from \texttt{conv5}.

Finally, Figure~\ref{fig:waouh} shows the top $9$ activated images of some \texttt{conv5} filters that seem to be semantically coherent.
The filters on the top row contain information about structures that highly corrolate with object classes.
The filters on the bottom row seem to trigger on style, like drawings or abstract shapes.


\subsection{Linear classification on activations}
\label{sec:linear}

Following Zhang~\etal~\cite{zhang2016split}, we train a linear classifier on top of different frozen convolutional layers.
This layer by layer comparison with supervised features exhibits where a convnet starts to be task specific, \ie specialized in object classification.
We report the results of this experiment on ImageNet and the Places dataset~\cite{zhou2014learning} in Table~\ref{tab:linear}.
We choose the hyperparameters by cross-validation on the training set.
On ImageNet, \OURS outperforms the state of the art from \texttt{conv3} to \texttt{conv5} layers by $3-5\%$.
The largest improvement is observed in the \texttt{conv4} layer, while the \texttt{conv1} layer performs poorly,
probably because the Sobel filtering discards color.
Consistently with the filter visualizations of Sec.~\ref{sec:viz}, \texttt{conv3} works better than \texttt{conv5}.
Finally, the difference of performance between \OURS and a supervised AlexNet grows significantly on higher layers:
at layers \texttt{conv2-conv3} the difference is only around $6\%$, 
but this difference rises to $14.4\%$ at \texttt{conv5}, marking where the AlexNet probably stores most of the class level information.
In the supplementary material, we also report the accuracy if a MLP is trained on the last layer; \OURS outperforms the state of the art by $8\%$.

\begin{table}[t]
  \centering
  \resizebox{\columnwidth}{!}{%
    \begin{tabular}{@{}l c ccccc c ccccc@{}}
      \toprule
            &~~~& \multicolumn{5}{c}{ImageNet} &~~~& \multicolumn{5}{c}{Places} \\
            \cmidrule{3-7} \cmidrule{9-13}
      Method && \texttt{conv1} & \texttt{conv2} & \texttt{conv3} & \texttt{conv4} & \texttt{conv5} && \texttt{conv1} & \texttt{conv2} & \texttt{conv3} & \texttt{conv4} & \texttt{conv5} \\
      \midrule
      Places labels                                     && --   & --   & --   & --   & --             && $22.1$ & $35.1$ & $40.2$ & $43.3$ & $44.6$  \\
      ImageNet labels                                   && $19.3$ & $36.3$ & $44.2$ & $48.3$ & $50.5$           && $22.7$ & $34.8$ & $38.4$ & $39.4$ & $38.7$  \\
      Random                                            && $11.6$ & $17.1$ & $16.9$ & $16.3$ & $14.1$           && $15.7$ & $20.3$ & $19.8$ & $19.1$ & $17.5$  \\
      \midrule
      Pathak~\etal~\cite{pathak2016context}             && $14.1$ & $20.7$ & $21.0$ & $19.8$ & $15.5$           && $18.2$ & $23.2$ & $23.4$ & $21.9$ & $18.4$  \\
      Doersch~\etal~\cite{doersch2015unsupervised}      && $16.2$ & $23.3$ & $30.2$ & $31.7$ & $29.6$           && $19.7$ & $26.7$ & $31.9$ & $32.7$ & $30.9$  \\
      Zhang~\etal~\cite{zhang2016colorful}              && $12.5$ & $24.5$ & $30.4$ & $31.5$ & $30.3$           && $16.0$ & $25.7$ & $29.6$ & $30.3$ & $29.7$  \\
      Donahue~\etal~\cite{donahue2016adversarial}       && $17.7$ & $24.5$ & $31.0$ & $29.9$ & $28.0$           && $21.4$ & $26.2$ & $27.1$ & $26.1$ & $24.0$  \\
      Noroozi and Favaro~\cite{noroozi2016unsupervised} && $\textbf{18.2}$ & $28.8$ & $34.0$ & $33.9$ & $27.1$  && $23.0$ & $32.1$ & $35.5$ & $34.8$ & $31.3$  \\
      Noroozi~\etal~\cite{noroozi2017representation}    && $18.0$ & $\textbf{30.6}$ & $34.3$ & $32.5$ & $25.7$           && $\textbf{23.3}$ & $\textbf{33.9}$ & $36.3$ & $34.7$ & $29.6$  \\
      Zhang~\etal~\cite{zhang2016split}                 && $17.7$ & $29.3$ & $35.4$ & $35.2$ & $32.8$           && $21.3$ & $30.7$ & $34.0$ & $34.1$ & $32.5$  \\
      \midrule
      \OURS  && $12.9$ & $29.2$ & $\textbf{38.2}$ & $\textbf{39.8}$ & $\textbf{36.1}$ && $18.6$ & $30.8$ & $\textbf{37.0}$ & $\textbf{37.5}$ & $\textbf{33.1}$ \\
      \bottomrule
    \end{tabular}
  }
  \vspace{\soustable}
  \caption{
    Linear classification on ImageNet and Places using activations from the convolutional layers of an AlexNet as features.
    We report classification accuracy on the central crop.
    Numbers for other methods are from Zhang~\etal~\cite{zhang2016split}.
  }
  \label{tab:linear}
\end{table}

The same experiment on the Places dataset provides some interesting insights:
like \OURS, a supervised model trained on ImageNet suffers from a decrease of performance for higher layers (\texttt{conv4} versus \texttt{conv5}).
Moreover, \OURS yields \texttt{conv3-4} features that are comparable to those trained with ImageNet labels.
This suggests that when the target task is sufficently far from the domain covered by ImageNet, labels are less important.


\subsection{\textsc{Pascal} VOC 2007}
\label{sec:pascal}
Finally, we do a quantitative evaluation of \OURS on image classification, object detection and semantic segmentation on \textsc{Pascal} VOC.
The relatively small size of the training sets on \textsc{Pascal} VOC ($2,500$ images) makes this setup closer to a ``real-world'' application,
where a model trained with heavy computational resources, is adapted to a task or a dataset with a small number of instances.
Detection results are obtained using \texttt{fast-rcnn}\footnote{https://github.com/rbgirshick/py-faster-rcnn}; segmentation results are obtained using the code of Shelhamer~\etal\footnote{https://github.com/shelhamer/fcn.berkeleyvision.org}.
For classification and detection, we report the performance on the test set of \textsc{Pascal} VOC 2007 and choose our hyperparameters on the validation set.
For semantic segmentation, following the related work, we report the performance on the validation set of \textsc{Pascal} VOC 2012.

\begin{table}[t]
  \centering
  \begin{tabular}{lc cc c cc c cc}
    \toprule
    &~~~~~~~~& \multicolumn{2}{c}{Classification} &~~~& \multicolumn{2}{c}{Detection} &~~~& \multicolumn{2}{c}{Segmentation} \\
                      \cmidrule{3-4} \cmidrule{6-7} \cmidrule{9-10}
    Method && \textsc{fc6-8}& \textsc{all} && \textsc{fc6-8}& \textsc{all} && \textsc{fc6-8} & \textsc{all} \\
    \midrule
    ImageNet labels                                             && $~78.9~$ & $~79.9~$ && --       & $~56.8~$ && --       & $~48.0~$ \\
    Random-rgb                                                  && $~33.2~$ & $~57.0~$ && $~22.2~$ & $~44.5~$ && $~15.2~$ & $~30.1~$ \\
    Random-sobel                                                && $~29.0~$ & $~61.9~$ && $~18.9~$ & $~47.9~$ && $~13.0~$ & $~32.0~$ \\
    \midrule
    Pathak~\etal~\cite{pathak2016context}                       && $~34.6~$ & $~56.5~$ && --   & $~44.5~$ && --   & $~29.7~$ \\
    Donahue~\etal~\cite{donahue2016adversarial}$^*$             && $~52.3~$ & $~60.1~$ && --   & $~46.9~$ && --   & $~35.2~$ \\
    Pathak~\etal~\cite{pathak2016learning}                      && --       & $~61.0~$ && --   & $~52.2~$ && --   & --     \\
		Owens~\etal~\cite{owens2016ambient}$^*$                     && $~52.3~$ & $~61.3~$ && --   & --   && --   & --   \\
    Wang and Gupta~\cite{wang2015unsupervised}$^*$              && $~55.6~$ & $~63.1~$ && $~32.8^\dagger$   & $~47.2~$ && $~26.0^\dagger$   & $~35.4^\dagger$   \\
    Doersch~\etal~\cite{doersch2015unsupervised}$^*$            && $~55.1~$ & $~65.3~$ && --   & $~51.1~$ && --   & --   \\
    Bojanowski and Joulin~\cite{bojanowski2017unsupervised}$^*$ && $~56.7~$ & $~65.3~$ && $~33.7^\dagger$ & $~49.4~$ && $~26.7^\dagger$ & $~37.1^\dagger$ \\ 
    Zhang~\etal~\cite{zhang2016colorful}$^*$                    && $~61.5~$ & $~65.9~$ && $~43.4^\dagger$ & $~46.9~$ && $~35.8^\dagger$ & $~35.6~$ \\ 
    Zhang~\etal~\cite{zhang2016split}$^*$                       && $~63.0~$ & $~67.1~$ && --   & $~46.7~$ && --   & $~36.0~$ \\
    Noroozi and Favaro~\cite{noroozi2016unsupervised}           && --       & $~67.6~$ && --   & $~53.2~$ && --   & $~37.6~$  \\
    Noroozi~\etal~\cite{noroozi2017representation}              && --       & $~67.7~$ && --   & $~51.4~$ && --   & $~36.6~$ \\
		\midrule
    \OURS && $\textbf{70.4}$ & $\textbf{73.7}$ && $\textbf{51.4}$ & $\textbf{55.4}$ && $\textbf{43.2}$ & $\textbf{45.1}$ \\
    \bottomrule
  \end{tabular}
  \vspace{\soustable}
  \caption{
    Comparison of the proposed approach to state-of-the-art unsupervised feature learning on classification, detection and segmentation on \textsc{Pascal} VOC.
    $^*$ indicates the use of the data-dependent initialization of Kr\"ahenb\"uhl~\etal~\cite{krahenbuhl2015data}.
    Numbers for other methods produced by us are marked with a $\dagger$.
  }
  \label{tab:voc}
\end{table}

Table~\ref{tab:voc} summarized the comparisons of \OURS with other feature-learning approaches on the three tasks.
As for the previous experiments, we outperform previous unsupervised methods on all three tasks, in every setting.
The improvement with fine-tuning over the state of the art is the largest on semantic segmentation ($7.5\%$).
On detection, \OURS performs only slightly better than previously published methods.
Interestingly, a fine-tuned random network performs comparatively to many unsupervised methods, but performs poorly if only \textsc{fc6-8} are learned.
For this reason, we also report detection and segmentation with \textsc{fc6-8} for \OURS and a few baselines.
These tasks are closer to a real application where fine-tuning is not possible.
It is in this setting that the gap between our approach and the state of the art is the greater (up to $9\%$ on classification).


%% file: discussion.tex


\section{Discussion}

The current standard for the evaluation of an unsupervised method involves the use of an AlexNet architecture trained on ImageNet and tested on class-level tasks.
To understand and measure the various biases introduced by this pipeline on \OURS, we consider a different training set, a different architecture and an instance-level recognition task.

\begin{table}[t!]
  \centering
  \begin{tabular}{@{}lc cc cc c cc c cc@{}}
    \toprule
                      &~~~~&            &~~~~& \multicolumn{2}{c}{Classification} &~~& \multicolumn{2}{c}{Detection} &~~& \multicolumn{2}{c}{Segmentation} \\
                      \cmidrule{5-6} \cmidrule{8-9} \cmidrule{11-12}
    Method    && Training set && \textsc{fc6-8}& \textsc{all}   && \textsc{fc6-8}& \textsc{all}  && \textsc{fc6-8}& \textsc{all} \\
                      \midrule
    Best competitor &&ImageNet&&  $~63.0~$ & $~67.7~$ && $~43.4^{\dagger}$ & $~53.2~$ && $~35.8^{\dagger}$ & $~37.7~$ \\
                      \midrule
    \OURS&& ImageNet   && $~72.0~$  & $~73.7~$  && $~51.4~$ & $~55.4~$ && $~43.2~$ & $~45.1~$ \\
    \OURS&& YFCC100M   && $~67.3~$  & $~69.3~$  && $~45.6~$ & $~53.0~$ && $~39.2~$ & $~42.2~$ \\
    \bottomrule
  \end{tabular}
  \vspace{\soustable}
  \caption{
    Impact of the training set on the performance of \OURS measured on the \textsc{Pascal} VOC transfer tasks as described in Sec.~\ref{sec:pascal}.
    We compare ImageNet with a subset of $1$M images from YFCC100M~\cite{thomee2015new}.
    Regardless of the training set, \OURS outperforms the best published numbers on most tasks.
  Numbers for other  methods produced by us are marked with a ${\dagger}$
  }
  \label{tab:flickr}
\end{table}

\subsection{ImageNet versus YFCC100M}

ImageNet is a dataset designed for a fine-grained object classification challenge~\cite{russakovsky2015imagenet}.
It is object oriented, manually annotated and organised into well balanced object categories.
By design, \OURS favors balanced clusters and, as discussed above, our number of cluster $k$ is somewhat comparable with the number of labels in ImageNet.
This may have given an unfair advantage to \OURS over other unsupervised approaches when trained on ImageNet.
To measure the impact of this effect, we consider a subset of randomly-selected $1$M images from the YFCC100M dataset~\cite{thomee2015new} for the pre-training.
Statistics on the hashtags used in YFCC100M suggests that the underlying ``object classes'' are severly unbalanced~\cite{joulin2016learning},
leading to a data distribution less favorable to \OURS.

Table~\ref{tab:flickr} shows the difference in performance on \textsc{Pascal} VOC of \OURS pre-trained on YFCC100M compared to ImageNet.
As noted by Doersch~\etal~\cite{doersch2015unsupervised}, this dataset is not object oriented, hence the performance are expected to drop by a few percents.
However, even when trained on uncured Flickr images, \OURS outperforms the current state of the art by a significant margin on most tasks (up to $+4.3\%$ on classification and $+4.5\%$ on semantic segmentation).
We report the rest of the results in the supplementary material with similar conclusions.
This experiment validates that \OURS is robust to a change of image distribution, leading
to state-of-the-art general-purpose visual features even if this distribution is not favorable to its design.

\subsection{AlexNet versus VGG}

In the supervised setting, deeper architectures like VGG or ResNet~\cite{he2015delving} have a much higher accuracy on ImageNet than AlexNet.
We should expect the same improvement if these architectures are used with an unsupervised approach.
Table~\ref{tab:vocVGG} compares a VGG-16 and an AlexNet trained with \OURS on ImageNet and tested on the \textsc{Pascal} VOC 2007 object detection task with fine-tuning.
We also report the numbers obtained with other unsupervised approaches~\cite{doersch2015unsupervised,wang2017transitive}.
Regardless of the approach, a deeper architecture leads to a significant improvement in performance on the target task.
Training the VGG-16 with \OURS gives a performance above the state of the art, bringing us to only $1.4$ percents below the supervised topline.
Note that the difference between unsupervised and supervised approaches remains in the same ballpark for both architectures (\ie $1.4\%$).
Finally, the gap with a random baseline grows for larger architectures,
justifying the relevance of unsupervised pre-training for complex architectures when little supervised data is available.

\begin{table}[t!]
  \parbox[c][16em][t]{.45\linewidth}{
    \centering
    \begin{tabular}{@{}l c c c c@{}}
      \toprule
      Method &~~~& AlexNet && VGG-16 \\
      \midrule
      ImageNet labels                 && $56.8$ && $67.3$ \\
      Random                          && $47.8$ && $39.7$ \\
      \midrule
      Doersch~\etal~\cite{doersch2015unsupervised}  && $51.1$ && $61.5$ \\
      Wang and Gupta~\cite{wang2015unsupervised}    && $47.2$ && $60.2$ \\
      Wang~\etal~\cite{wang2017transitive}          && --   && $63.2$ \\
      \midrule
      \OURS                           && $\textbf{55.4}$ && $\textbf{65.9}$ \\
      \bottomrule
    \end{tabular}
    \vspace{\soustable}
    \caption{
      \textsc{Pascal} VOC 2007 object detection with AlexNet and VGG-16.
      Numbers are taken from Wang~\etal~\cite{wang2017transitive}.
    }
    \label{tab:vocVGG}
  }
  \hfill
  \parbox[c][16em][t]{.45\linewidth}{
    \centering
    \begin{tabular}{@{}lc c c c@{}}
      \toprule
      Method &~~~& Oxford$5$K && Paris$6$K \\
      \midrule
      ImageNet labels              && $72.4$ && $81.5$  \\
      Random                       && $~6.9$  && $22.0$ \\
      \midrule
      Doersch~\etal~\cite{doersch2015unsupervised}  && $35.4$ && $53.1$ \\
      Wang~\etal~\cite{wang2017transitive}          && $42.3$ && $58.0$ \\
      \midrule
      \OURS                           && $\textbf{61.0}$  && $\textbf{72.0}$ \\
      \bottomrule
    \end{tabular}
    \vspace{\soustable}
    \caption{
      mAP on instance-level image retrieval on Oxford and Paris dataset with a VGG-16.
      We apply R-MAC with a resolution of $1024$ pixels and $3$ grid levels~\cite{tolias2015particular}.
    }
    \label{tab:retrieval}
  }
\end{table}

\subsection{Evaluation on instance retrieval}

The previous benchmarks measure the capability of an unsupervised network to capture class level information.
They do not evaluate if it can differentiate images at the instance level.
To that end, we propose image retrieval as a down-stream task.
We follow the experimental protocol of Tolias~\etal~\cite{tolias2015particular} on two datasets, \ie, Oxford Buildings~\cite{Philbin07} and Paris~\cite{Philbin08}.
Table~\ref{tab:retrieval} reports the performance of a VGG-16 trained with different approaches obtained with Sobel filtering, except for Doersch~\etal~\cite{doersch2015unsupervised} and Wang~\etal~\cite{wang2017transitive}.
This preprocessing improves by $5.5$ points the mAP of a supervised VGG-16 on the Oxford dataset, but not on Paris.
This may translate in a similar advantage for \OURS, but it does not account for the average differences of $19$ points.
Interestingly, random convnets perform particularly poorly on this task compared to pre-trained models.
This suggests that image retrieval is a task where the pre-training is essential and studying it as a down-stream task could give further insights about the quality of the features produced by unsupervised approaches.

%% file: supp.tex
\title{Appendix}
\author{}
\institute{}
\maketitle
\section{Additional results}

\subsection{Classification on ImageNet}
\label{sec:mlpImageNet}
Noroozi and Favaro~\cite{noroozi2016unsupervised} suggest to evaluate networks trained in an unsupervised way by freezing the convolutional layers and retrain on ImageNet the fully connected layers using labels and reporting accuracy on the validation set.
This experiment follows the observation of Yosinki \etal~\cite{yosinski2014transferable} that general-purpose features appear in the convolutional layers of an AlexNet.
We report a comparison of \OURS to other AlexNet networks trained with no supervision, as well as random and supervised baselines in Table~\ref{tab:mlpImageNet}.

\begin{table}[t]
  \centering
  \begin{tabular}{@{}lcr@{}}
    \toprule
    Method & Pre-trained dataset & Acc@1 \\
    \midrule
    Supervised & ImageNet  & 59.7 \\
    Supervised Sobel & ImageNet  & 57.8 \\
    Random & - & 12.0 \\
    \midrule
    Wang \etal~\cite{wang2015unsupervised} & YouTube100K~\cite{LiangLWLLY14} & 29.8 \\
    \midrule
    Doersch \etal~\cite{doersch2015unsupervised} & ImageNet & 30.4 \\
    Donahue \etal~\cite{donahue2016adversarial}  & ImageNet & 32.2 \\
    Noroozi and Favaro~\cite{noroozi2016unsupervised} & ImageNet & 34.6 \\
    Zhang \etal~\cite{zhang2016colorful} & ImageNet & 35.2 \\
    Bojanowski and Joulin~\cite{bojanowski2017unsupervised} & ImageNet & 36.0 \\
    \midrule
    \OURS & ImageNet & 44.0 \\
    \OURS & YFCC100M & 39.6 \\
    \bottomrule
  \end{tabular}
  \caption{
    Comparison of \OURS to AlexNet features pre-trained supervisedly and unsupervisedly on different datasets.
    A full multi-layer perceptron is retrained on top of the frozen pre-trained features.
    We report classification accuracy (acc@1).
    Expect for Noroozi and Favaro~\cite{noroozi2016unsupervised},
    all the numbers are taken from Bojanowski and Joulin~\cite{bojanowski2017unsupervised}.
  }
  \label{tab:mlpImageNet}
\end{table}

\OURS outperforms state-of-the-art unsupervised methods by a significant margin, achieving $8\%$ better accuracy than the previous best performing method.
This means that \OURS \emph{halves the gap} with networks trained in a supervised setting.

\subsection{Stopping criterion}
\begin{figure}[!h]
\centering
\includegraphics[scale = 0.5]{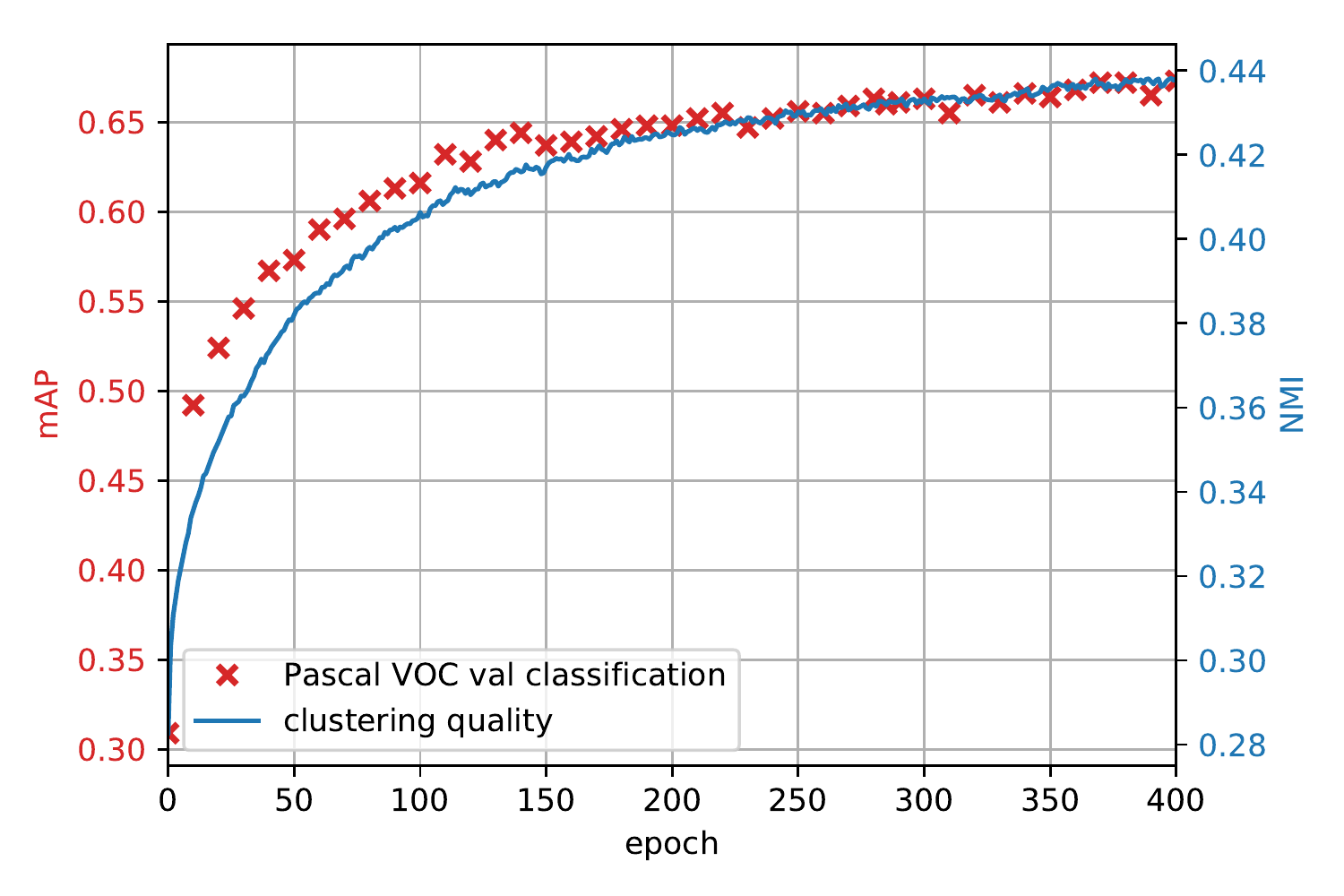}
\caption{In red: validation mAP \textsc{Pascal} VOC classification performance.
         In blue: evolution of the clustering quality.}
\label{convergence}
\end{figure}

We monitor how the features learned with \OURS evolve along the training epochs on a down-stream task: object classification on the validation set of \textsc{Pascal} VOC with no fine-tuning.
We use this measure to select the hyperparameters of our model as well as to check when the features stop improving.
In Figure~\ref{convergence}, we show the evolution of both the classification accuracy on this task and a measure of the clustering quality (NMI between the cluster assignments and the true labels) throughout the training.
Unsurprisingly, we notice that the clustering and features qualities follow the same dynamic.
The performance saturates after $400$ epochs of training.

\section{Further discussion}

In this section, we discuss some technical choices and variants of \OURS more specifically.

\subsection{Alternative clustering algorithm}

\noindent\textbf{Graph clustering.}
We consider Power Iteration Clustering (PIC)~\cite{lin2010power} as an alternative clustering method.
It has been shown to yield good performance for large scale collections~\cite{douze2017evaluation}.

Since PIC is a graph clustering approach, we generate a nearest neighbor graph by connecting all images to their 5 neighbors in the Euclidean space of image descriptors. 
We denote by $f_\theta(x)$ the output of the network with parameters $\theta$ applied to image $x$. 
We use the sparse graph matrix $G=\mathbb{R}^{n \times n}$. 
We set the diagonal of $G$ to 0 and non-zero entries are defined as 
\[
w_{ij} = e^{-\frac{\|f_\theta(x_i) - f_\theta(x_j)\|^2}{\sigma^2}}
\]
with $\sigma$ a bandwidth parameter.
In this work, we use a variant of PIC~\cite{douze2017evaluation,CL12} that does: 
\begin{enumerate}
\item 
	Initialize $v \leftarrow [1/n,..., 1/n]^\top \in \mathbb{R}^n$;
\item  
	Iterate \[
	v \leftarrow N_1(\alpha (G + G^\top) v + (1 - \alpha) v),
	\]
	where $\alpha = 10^{-3}$ is a regularization parameter and $N_1:v \mapsto v / \|v\|_1$ the L1-normalization function; 
\item 
	Let $G'$ be the directed unweighted subgraph of $G$ where we keep edge $i \rightarrow j$ of $G$ such that 
	\[
	j = \mathrm{argmax}_j w_{ij}(v_j - v_i). 
	\] 
	If $v_i$ is a local maximum (ie. $\forall j\ne i, v_j \le v_i$), then no edge starts from it. 
	The clusters are given by the connected components of $G'$. 
	Each cluster has one local maximum. 
\end{enumerate}

An advantage of PIC clustering is not to require the setting beforehand of the number of clusters.
However, the parameter $\sigma$ influences the number of clusters: when it is larger, the edges become more uniform and the number of clusters decreases, and the other way round when $\sigma$ increased. 
In the following, we set $\sigma=0.2$.

\begin{figure}[!h]
\centering

\begin{tabular}{ccccccc}
Query &&& Random &&& \OURS PIC
\\
\includegraphics[scale = 0.2]{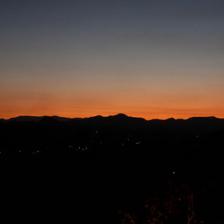}&&&
\includegraphics[scale = 0.2]{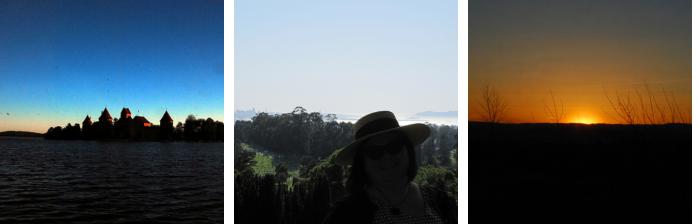}&&&
\includegraphics[scale = 0.2]{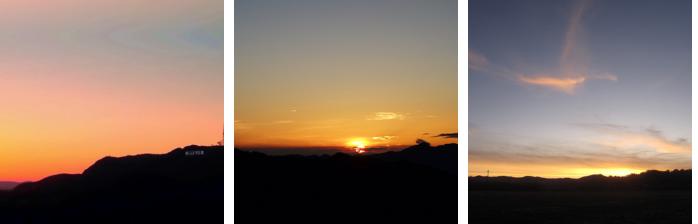}
\\

 \includegraphics[scale = 0.2]{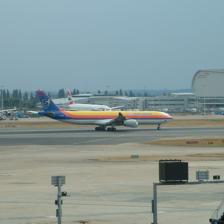}&&&
 \includegraphics[scale = 0.2]{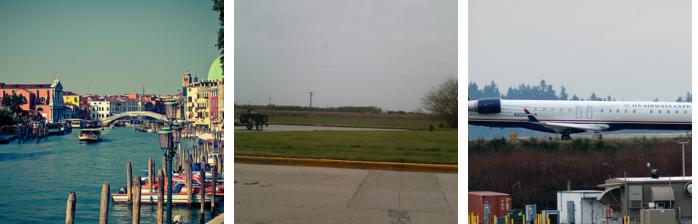}&&&
 \includegraphics[scale = 0.2]{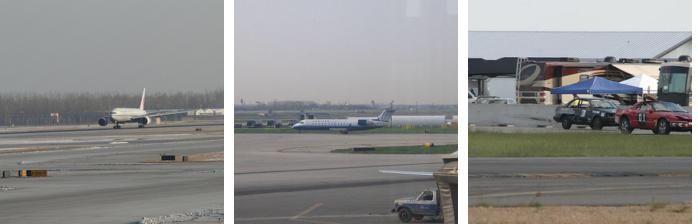}
\\
 \includegraphics[scale = 0.2]{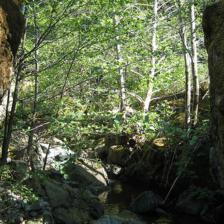}&&&
 \includegraphics[scale = 0.2]{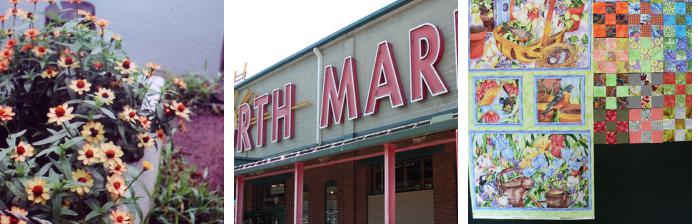}&&&
 \includegraphics[scale = 0.2]{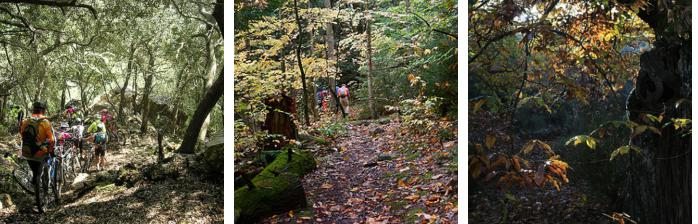}
\\

\includegraphics[scale = 0.2]{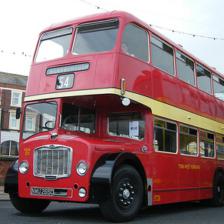}&&&
\includegraphics[scale = 0.2]{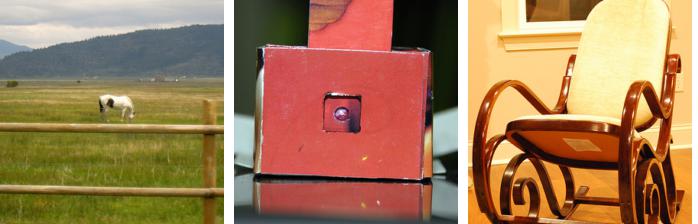}&&&
\includegraphics[scale = 0.2]{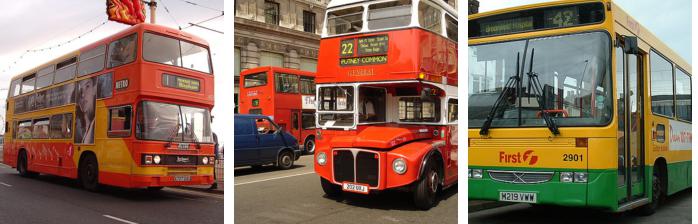}
\end{tabular}

\caption{Images and their $3$ nearest neighbors in a subset of Flickr in the feature space. The query images are shown on the left column.
 The following $3$ columns correspond to a randomly initialized network and the last $3$ to the same network after training with PIC \OURS.
}
\label{nn-retrieval}
\end{figure}

As our PIC implementation relies on a graph of nearest neighbors, we show in Figure~\ref{nn-retrieval} some query images and their $3$ nearest neighbors in the feature space with a network trained with the PIC version of DeepCluster and a random baseline.
A randomly initialized network performs quite well in some cases (sunset query for example), where the image has a simple low-level structure.
The top row in Fig.~\ref{nn-retrieval} seems to represent a query for which the performance of the random network is quite good whereas the bottom query is too complex for the random network to retrieve good matches.
Moreover, we notice that the nearest neighbors matches are largely improved after the network has been trained with \OURS.

\begin{figure}[!h]
\centering
\includegraphics[scale = 0.5]{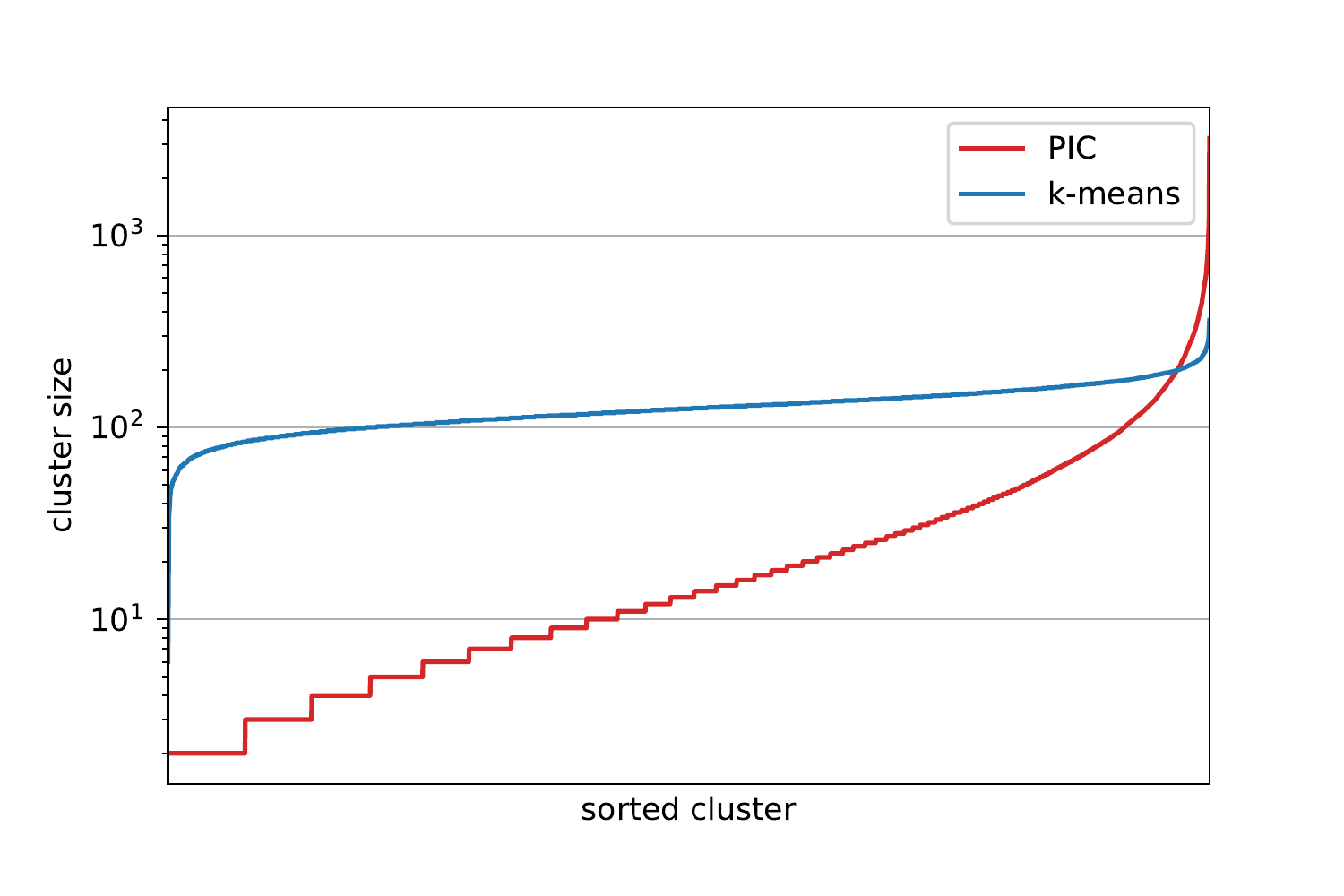}
\caption{Sizes of clusters produced by the $k$-means and PIC versions of \OURS at the last epoch of training.}
\label{clustersize}
\end{figure}

\noindent\textbf{Comparison with $k$-means.}
First, we give an insight about the distribution of the images in the clusters. 
We show in Figure~\ref{clustersize} the sizes of the clusters produced by the $k$-means and PIC versions of \OURS at the last epoch of training (this distribution is stable along the epochs).
We observe that $k$-means produces more balanced clusters than PIC.
Indeed, for PIC,  almost one third of the clusters are of a size lower than $10$ while the biggest cluster contains roughly $3000$ examples.
In this situation of very unbalanced clusters, it is important in our method to train the convnet by sampling images based on a uniform distribution over the clusters to prevent the biggest cluster from dominating the training.

We report in Table~\ref{tab:pic} the results for the different \textsc{Pascal} VOC transfer tasks with a model trained with the PIC version of DeepCluster.
For this set of transfer tasks, the models trained with $k$-means and PIC versions of \OURS perform in comparable ranges.

\begin{table}[h!]
  \centering
  \begin{tabular}{@{}l c cc c cc c cc@{}}
    \toprule
                      &Clustering algorithm& \multicolumn{2}{c}{Classification} &~~& \multicolumn{2}{c}{Detection} &~~& \multicolumn{2}{c}{Segmentation} \\
                      \cmidrule{3-4} \cmidrule{6-7} \cmidrule{9-10}
        && \textsc{fc6-8} & \textsc{all}   && \textsc{fc6-8} & \textsc{all}  && \textsc{fc6-8} & \textsc{all} \\
                      \midrule
    \OURS& $k$-means  & 72.0  & 73.7  && 51.4 & 55.4 && 43.2 & 45.1 \\
    \OURS& PIC        & 71.0  & 73.0  && 53.6 & 54.4 && 42.4 & 43.8 \\
    \bottomrule
  \end{tabular}
  \caption{
    Evaluation of PIC versus $k$-means on \textsc{Pascal} VOC transfer tasks.
  }
  \label{tab:pic}
\end{table}

\subsection{Variants of \OURS}
In Table~\ref{tab:flick},  we report the results of models trained with different variants of \OURS: a different training set, an alternative clustering method or without input preprocessing.
In particular, we notice that the performance of \OURS on raw RGB images degrades significantly.
\begin{table}[t]
  \centering
  \resizebox{\columnwidth}{!}{%
    \begin{tabular}{@{}l c ccccc c ccccc@{}}
      \toprule
            &~~~& \multicolumn{5}{c}{ImageNet} &~~~& \multicolumn{5}{c}{Places} \\
            \cmidrule{3-7} \cmidrule{9-13}
      Method && \texttt{conv1} & \texttt{conv2} & \texttt{conv3} & \texttt{conv4} & \texttt{conv5} && \texttt{conv1} & \texttt{conv2} & \texttt{conv3} & \texttt{conv4} & \texttt{conv5} \\
      \midrule
      Places labels                                     && --   & --   & --   & --   & --             && $22.1$ & $35.1$ & $40.2$ & $43.3$ & $44.6$  \\
      ImageNet labels                                   && $19.3$ & $36.3$ & $44.2$ & $48.3$ & $50.5$           && $22.7$ & $34.8$ & $38.4$ & $39.4$ & $38.7$  \\
      Random                                            && $11.6$ & $17.1$ & $16.9$ & $16.3$ & $14.1$           && $15.7$ & $20.3$ & $19.8$ & $19.1$ & $17.5$  \\
      \midrule
    Best competitor ImageNet && $\textbf{18.2}$ & $30.6$ & $35.4$ & $35.2$ & $32.8$ && $23.3$ & $\textbf{33.9}$ & $36.3$ & $34.7$ & $32.5$ \\
    \midrule
    \OURS            && $13.4$ & $32.3$ & $\textbf{41.0}$ & $39.6$ & $\textbf{38.2}$ && $19.6$ & $33.2$ & $\textbf{39.2}$ & $\textbf{39.8}$ & $34.7$ \\
    \OURS YFCC100M   && $13.5$ & $30.9$ & $38.0$ & $34.5$ & $31.4$ && $19.7$ & $33.0$ & $38.4$ & $39.0$ & $35.2$\\
    \OURS PIC        && $13.5$ & $32.4$ & $40.8$ & $\textbf{40.5}$ & $37.8$ && $19.5$ & $32.9$ & $39.1$ & $39.5$ & $\textbf{35.9}$ \\
    \OURS RGB        && $18.0$ & $\textbf{32.5}$ & $39.2$ & $37.2$ & $30.6$ && $\textbf{23.8}$ & $32.8$ & $37.3$ & $36.0$ & $31.0$ \\
    \bottomrule
  \end{tabular}
}
  \caption{
    Impact of different variants of our method on the performance of \OURS.
    We report classification accuracy of linear classifiers on ImageNet and Places using activations from the convolutional layers of an AlexNet as features.
    We compare the standard version of \OURS to different settings: YFCC100M as pre-training set; PIC as clustering method; raw RGB inputs.
  }
  \label{tab:flick}
\end{table}

In Table~\ref{tab:classif}, we compare the performance of \OURS depending on the clustering method and the pre-training set.
We evaluate this performance on three different classification tasks: with fine-tuning on \textsc{Pascal} VOC, without, and by retraining the full MLP on ImageNet.
We report the classification accuracy on the validation set.
Overall, we notice that the regular version of \OURS (with $k$-means) yields better results than the PIC variant on both ImageNet and the uncured dataset YFCC100M.

\begin{table}[h!]
  \centering
  \begin{tabular}{@{}lcccccc@{}}
    \toprule
     Dataset & Clustering&& \multicolumn{2}{c}{\textsc{Pascal} VOC} &~~& ImageNet \\
                                        \cmidrule{4-5} \cmidrule{7-7}
        &&& \textsc{fc6-8} & \textsc{all} && \textsc{fc6-8} \\
    \midrule
    ImageNet & $k$-means && 72.0  & 73.7  && 44.0  \\
    ImageNet & PIC        && 71.0  & 73.0  && 45.9 \\
    YFCC100M & $k$-means  && 67.3  & 69.3  && 39.6 \\
    YFCC100M & PIC        && 66.0  & 69.0  && 38.6 \\
    \bottomrule
  \end{tabular}
  \caption{
    Performance of \OURS with different pre-training sets and clustering algorithms measured as classification accuracy on \textsc{Pascal} VOC and ImageNet.
  }
  \label{tab:classif}
\end{table}

\section{Additional visualisation}
\subsection{Visualise VGG-$16$ features}
\begin{figure}[t]
\centering
\begin{tabular}{cccccccc}
\includegraphics[width=0.16\linewidth]{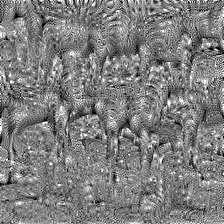}
\includegraphics[width=0.16\linewidth]{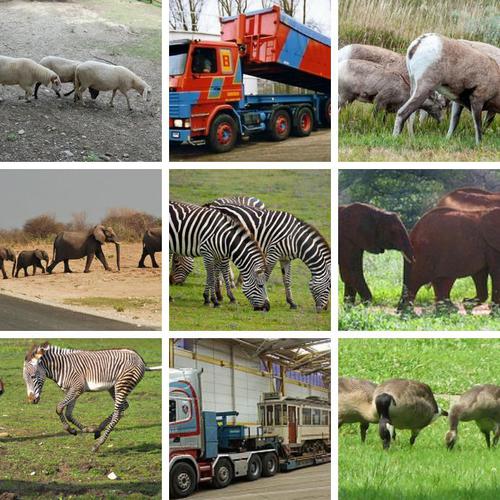}&
\includegraphics[width=0.16\linewidth]{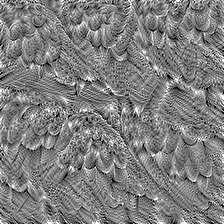}
\includegraphics[width=0.16\linewidth]{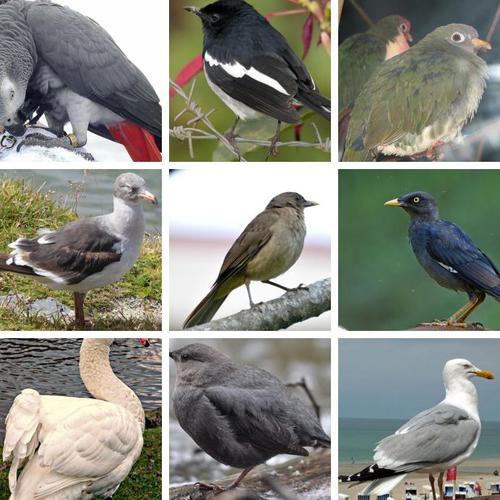}&
\includegraphics[width=0.16\linewidth]{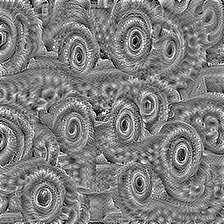}
\includegraphics[width=0.16\linewidth]{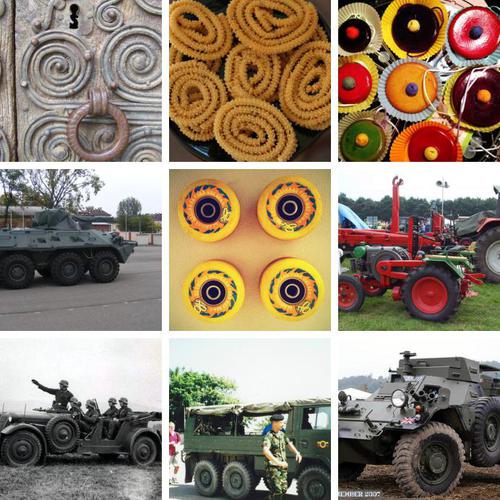}&
\\
\includegraphics[width=0.16\linewidth]{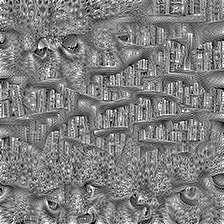}
\includegraphics[width=0.16\linewidth]{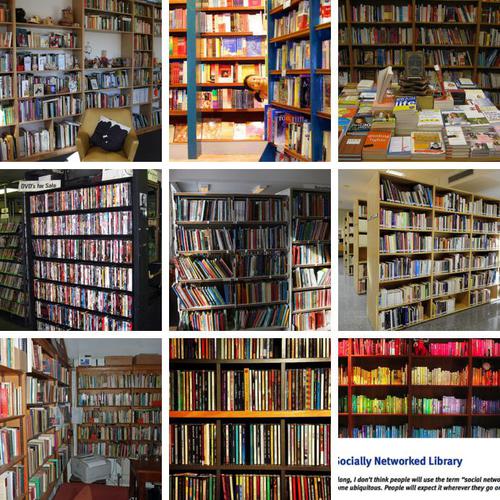}&
\includegraphics[width=0.16\linewidth]{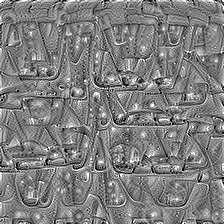}
\includegraphics[width=0.16\linewidth]{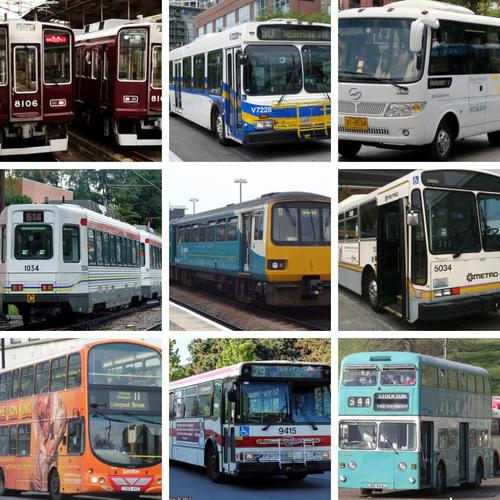}&
\includegraphics[width=0.16\linewidth]{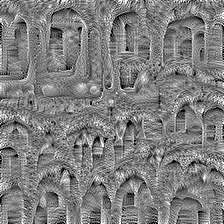}
\includegraphics[width=0.16\linewidth]{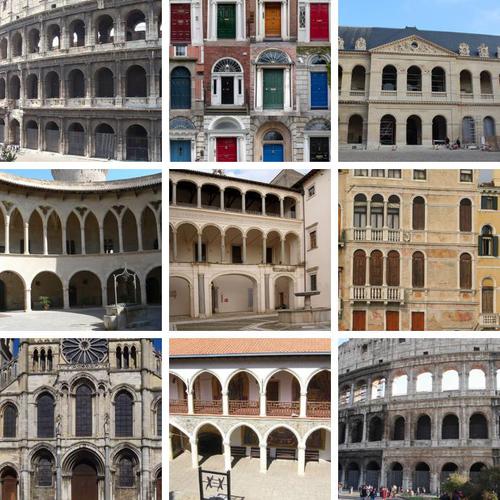}&
\\
\includegraphics[width=0.16\linewidth]{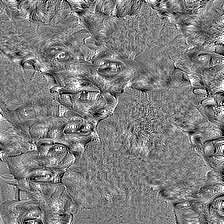}
\includegraphics[width=0.16\linewidth]{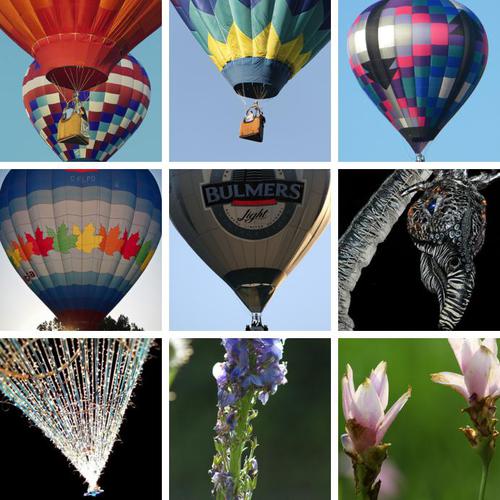}&
\includegraphics[width=0.16\linewidth]{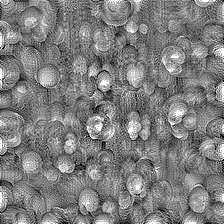}
\includegraphics[width=0.16\linewidth]{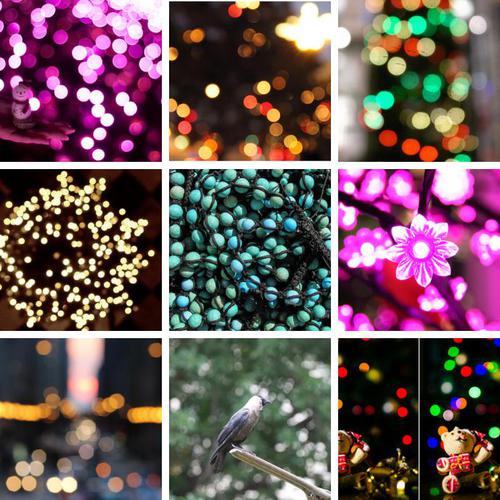}&
\includegraphics[width=0.16\linewidth]{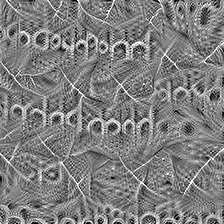}
\includegraphics[width=0.16\linewidth]{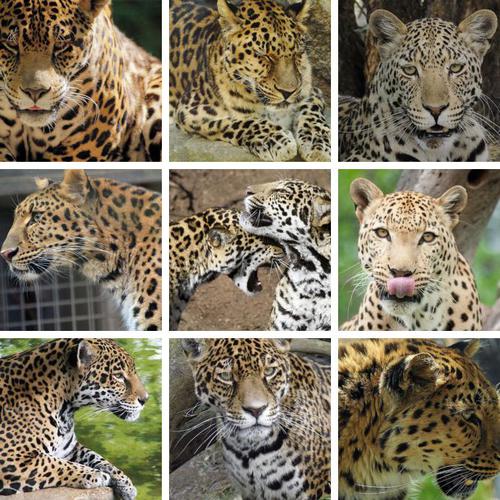}&
\\
\includegraphics[width=0.16\linewidth]{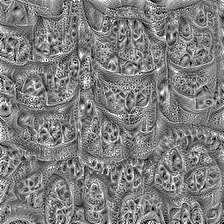}
\includegraphics[width=0.16\linewidth]{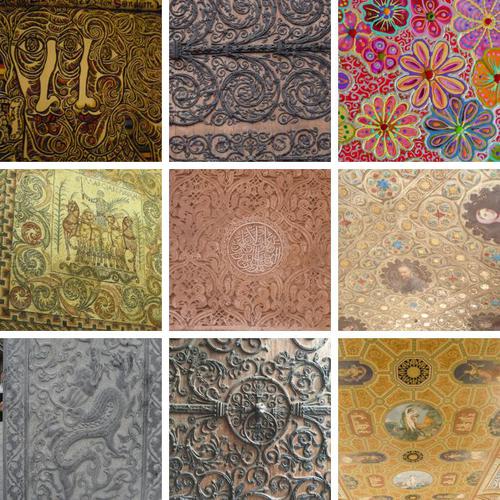}&
\includegraphics[width=0.16\linewidth]{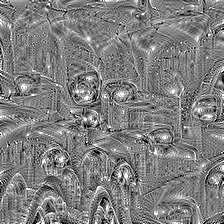}
\includegraphics[width=0.16\linewidth]{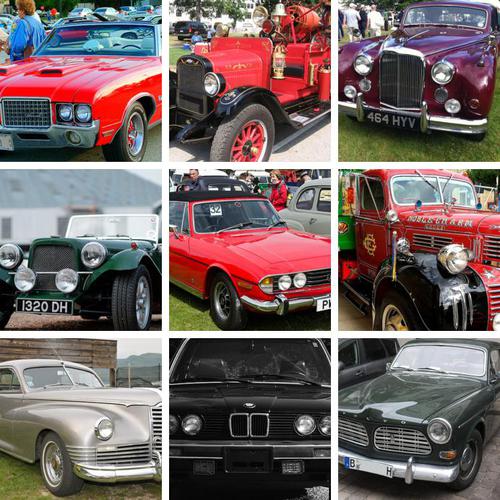}&
\includegraphics[width=0.16\linewidth]{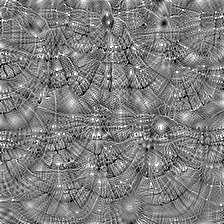}
\includegraphics[width=0.16\linewidth]{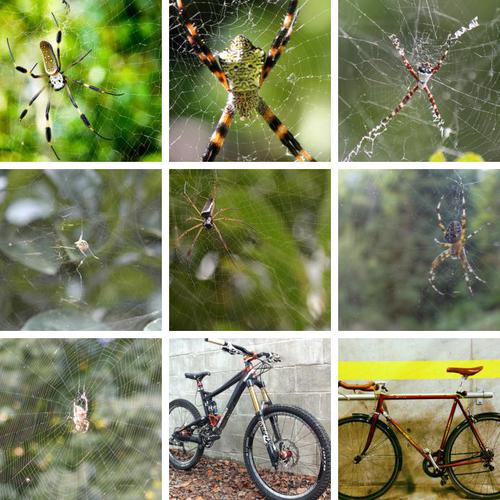}&

\end{tabular}
\caption{
Filter visualization and top 9 activated images (immediate right to the corresponding synthetic image) from a subset of 1 million images from YFCC100M for target filters in the last convolutional layer of a VGG-$16$ trained with \OURS.}
\label{fig:vgg}
\end{figure}

\begin{figure}[t]
\centering
\begin{tabular}{cccccccc}
\includegraphics[width=0.24\linewidth]{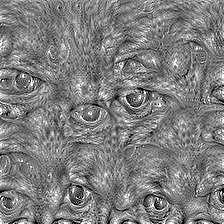}
\includegraphics[width=0.24\linewidth]{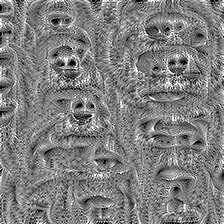}
\includegraphics[width=0.24\linewidth]{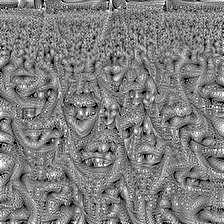}
\includegraphics[width=0.24\linewidth]{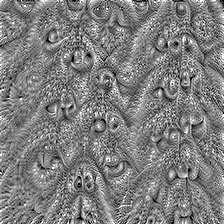}
\\
\includegraphics[width=0.24\linewidth]{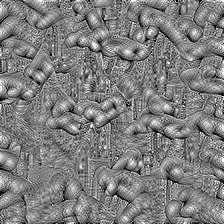}
\includegraphics[width=0.24\linewidth]{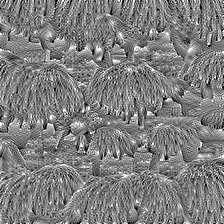}
\includegraphics[width=0.24\linewidth]{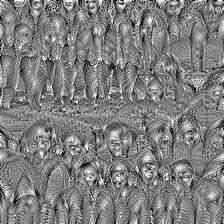}
\includegraphics[width=0.24\linewidth]{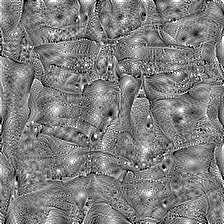}

\end{tabular}
\caption{
    Filter visualization by learning an input image that maximizes the response to a target filter~\cite{yosinski2015understanding} in the last convolutional layer of a VGG-$16$ convnet trained with \OURS.
    Here, we manually select filters that seem to trigger on human characteristics (eyes, noses, faces, fingers, fringes, groups of people or arms).
}
\label{fig:human}
\end{figure}

We assess the quality of the representations learned by the VGG-$16$ convnet with \OURS.
To do so, we learn an input image that maximises the mean activation of a target filter in the last convolutional layer.
We also display the top $9$ images in a random subset of $1$ million images from Flickr that activate the target filter maximally.
In Figure~\ref{fig:vgg}, we show some filters for which the top $9$ images seem to be semantically or stylistically coherent.
We observe that the filters, learned without any supervision, capture quite complex structures.
In particular, in Figure~\ref{fig:human}, we display synthetic images that correspond to filters that seem to focus on human characteristics.

\subsection{AlexNet}
\begin{figure}[t]
\centering
\includegraphics[width=1\linewidth]{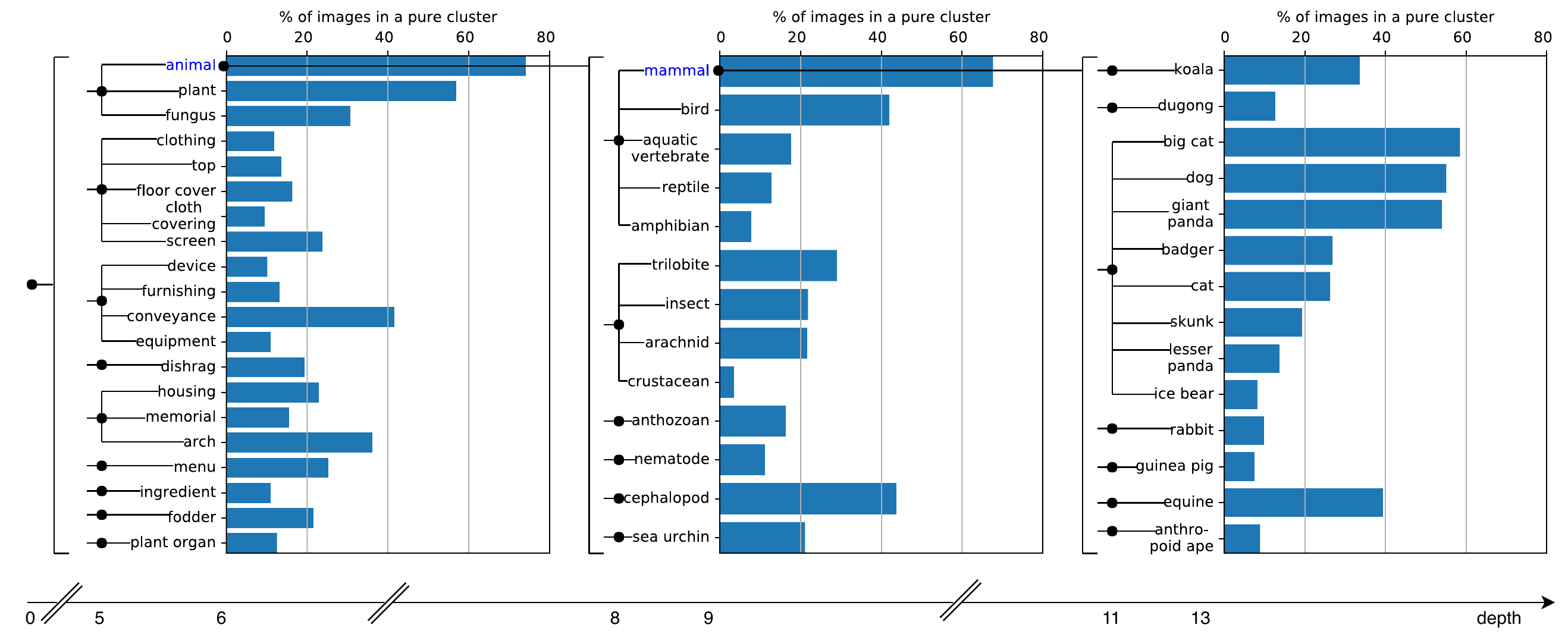} \\
\caption{
  {
    \footnotesize
    Proportion of images belonging to a \emph{pure cluster} for different synsets in the ImageNet ontology (pure cluster = more than 70\% of images are from that synset).
    We show synsets at depth 6, 9 and 13 in the ontology, with ``zooms'' on the animal and mammal synsets.
  }
}
\vspace{-1em}
\label{fig:clusters}
\end{figure}

It is interesting to investigate what clusters the unsupervised learning approach actually learns. 
Fig.~2(a) in the paper suggests that our clusters are correlated with ImageNet classes.
In Figure~\ref{fig:clusters}, we look at the purity of the clusters with the ImageNet ontology to see which concepts are learned.
More precisely, we show the proportion of images belonging to a \emph{pure cluster} for different synsets at different depths in the ImageNet ontology (pure cluster = more than 70\% of images are from that synset).
The correlation varies significantly between categories, with the highest correlations for animals and plants.

In Figure~\ref{fig:waouh2}, we display the top 9 activated images from a random subset of 10 millions images from YFCC100M for the first $100$ target filters in the last convolutional layer 
(we selected filters whose top activated images do not depict humans).
We observe that, even though our method is purely unsupervised, some filters trigger on images containing particular classes of objects.
Some other filters respond strongly on particular stylistic effects or textures.

\begin{figure}[t]
\centering
\begin{tabular}{ccccccc}
Filter $0$ && Filter $1$ && Filter $2$ && Filter $3$
\\
\includegraphics[width=0.23\linewidth]{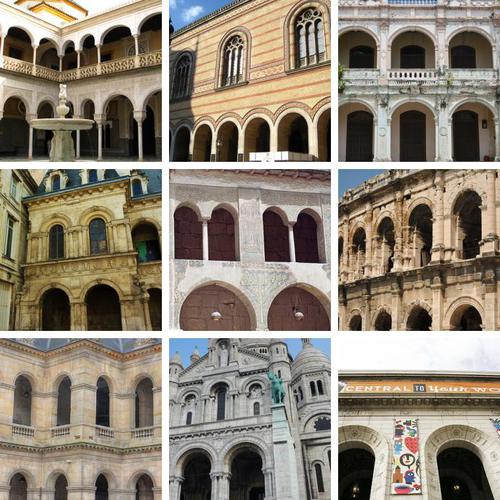}&&
\includegraphics[width=0.23\linewidth]{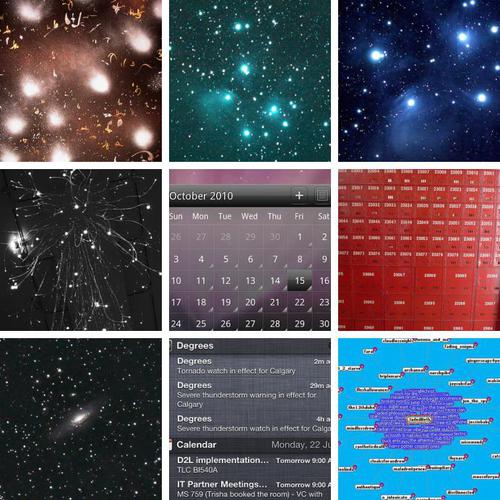}&&
\includegraphics[width=0.23\linewidth]{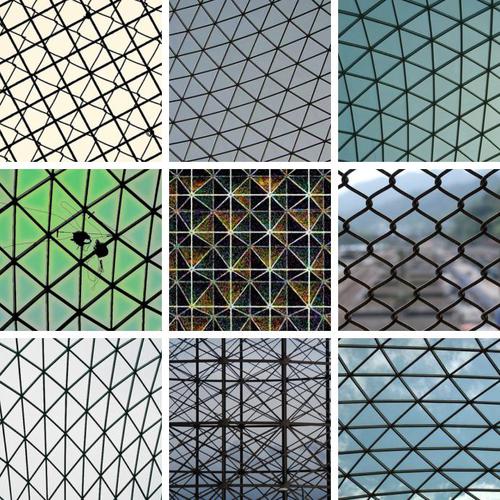}&&
\includegraphics[width=0.23\linewidth]{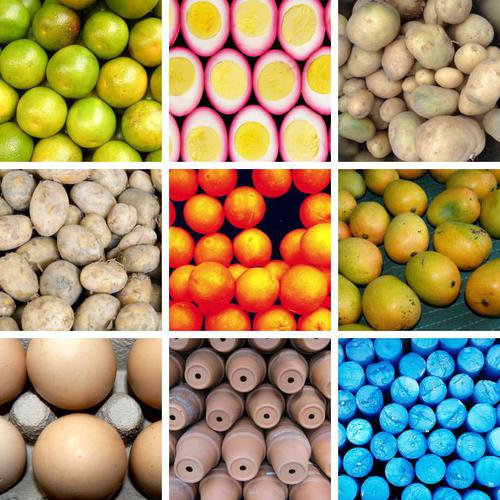}
\\
Filter $5$ && Filter $6$ && Filter $7$ && Filter $8$
\\
\includegraphics[width=0.23\linewidth]{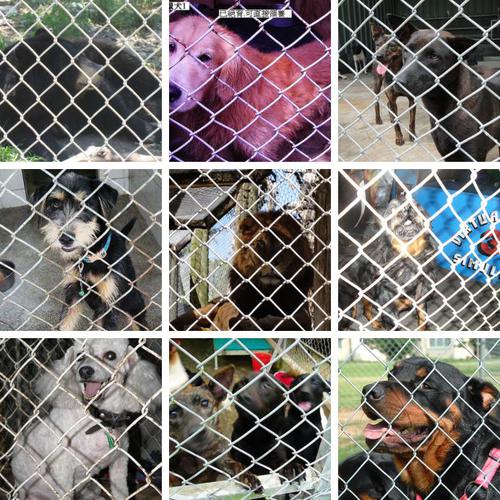}&&
\includegraphics[width=0.23\linewidth]{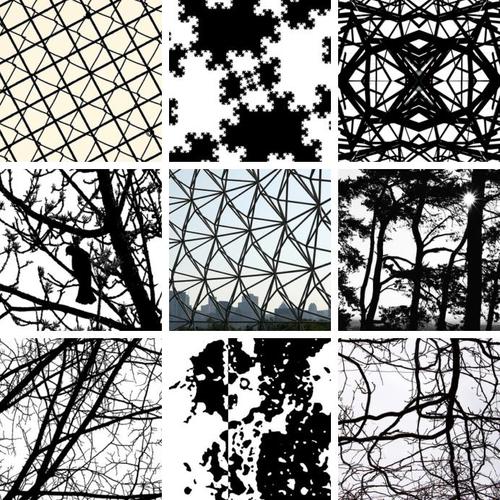}&&
\includegraphics[width=0.23\linewidth]{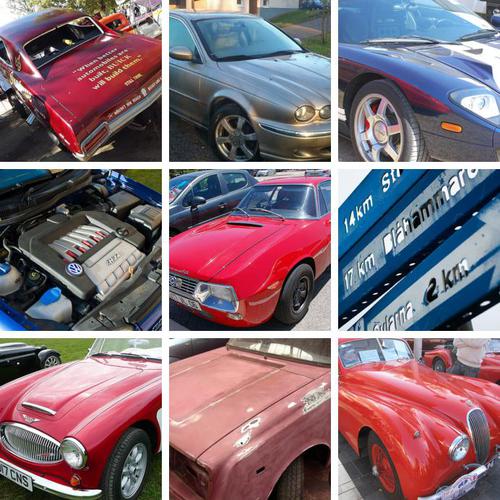}&&
\includegraphics[width=0.23\linewidth]{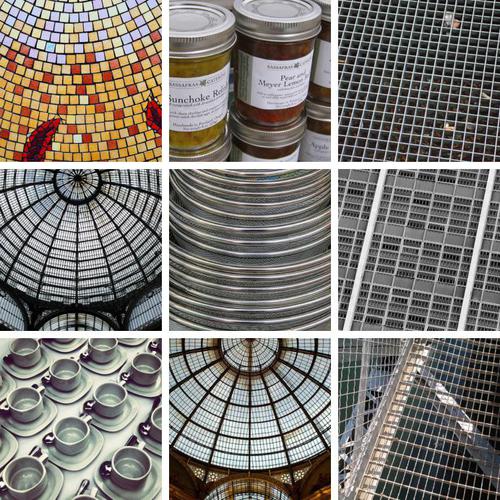}
\\
Filter $9$ && Filter $10$ && Filter $11$ && Filter $13$
\\
\includegraphics[width=0.23\linewidth]{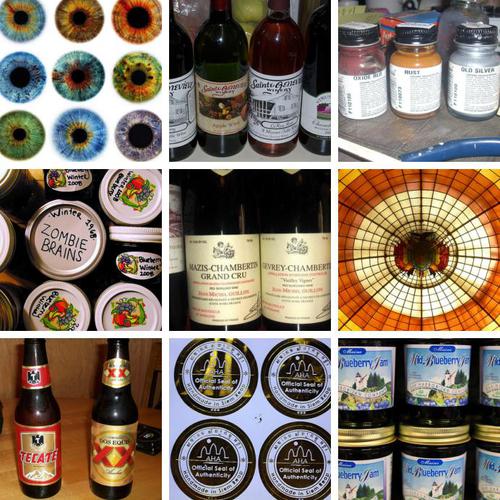}&&
\includegraphics[width=0.23\linewidth]{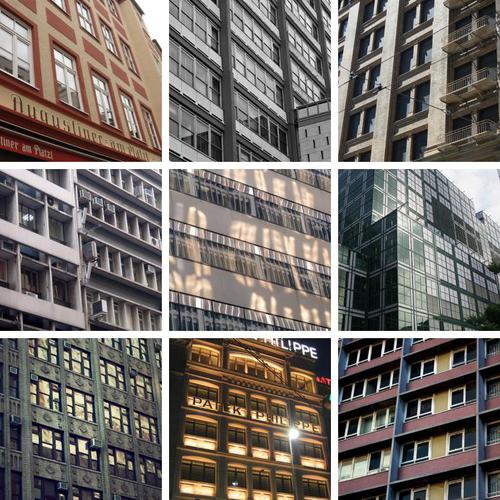}&&
\includegraphics[width=0.23\linewidth]{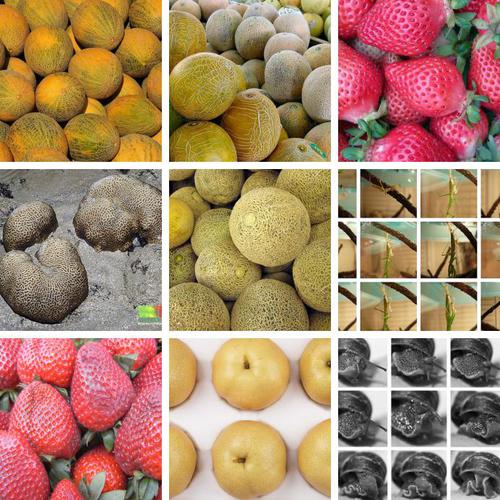}&&
\includegraphics[width=0.23\linewidth]{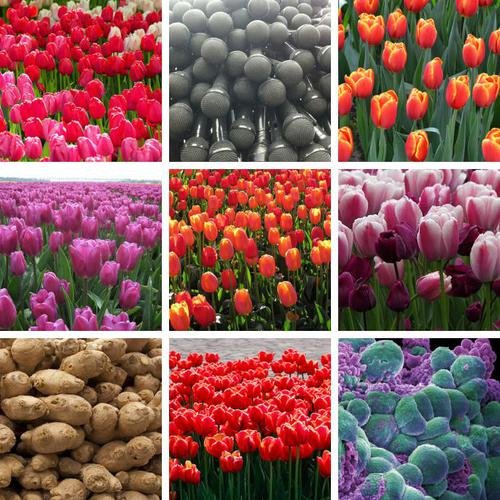}
\\
Filter $14$ && Filter $16$ && Filter $18$ && Filter $24$
\\
\includegraphics[width=0.23\linewidth]{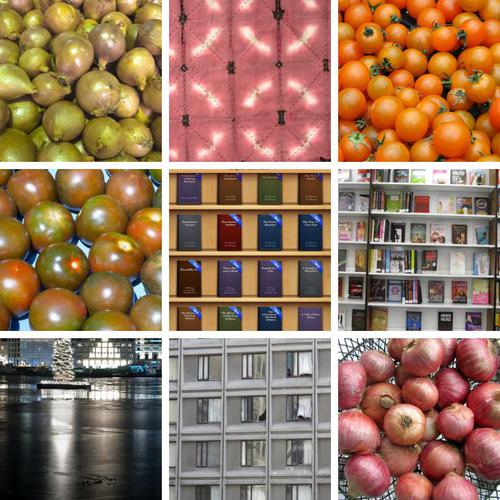}&&
\includegraphics[width=0.23\linewidth]{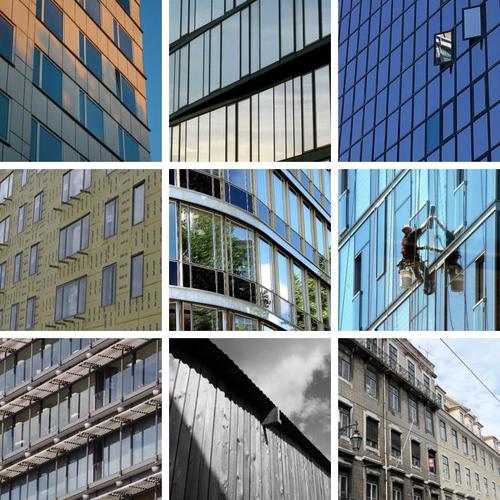}&&
\includegraphics[width=0.23\linewidth]{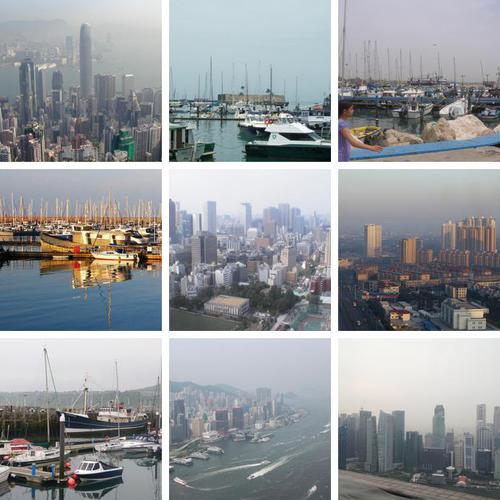}&&
\includegraphics[width=0.23\linewidth]{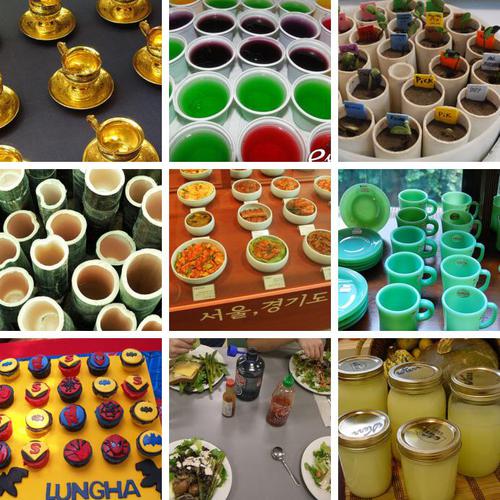}
\\
Filter $24$ && Filter $25$ && Filter $26$ && Filter $27$
\\
\includegraphics[width=0.23\linewidth]{images-filter-layer4-channel24-small.jpeg}&&
\includegraphics[width=0.23\linewidth]{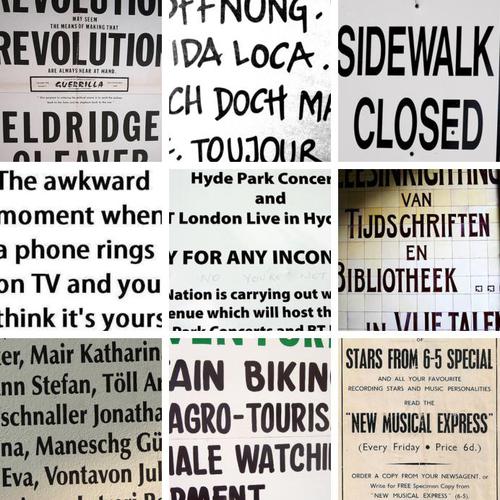}&&
\includegraphics[width=0.23\linewidth]{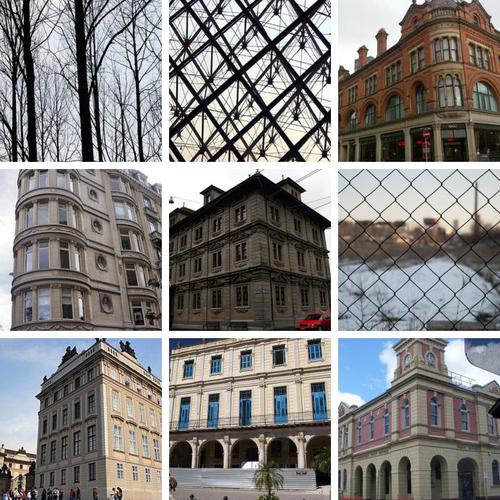}&&
\includegraphics[width=0.23\linewidth]{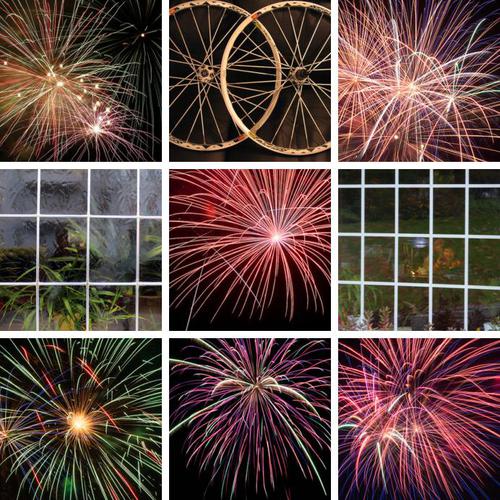}
\\
Filter $28$ && Filter $30$ && Filter $33$ && Filter $34$
\\
\includegraphics[width=0.23\linewidth]{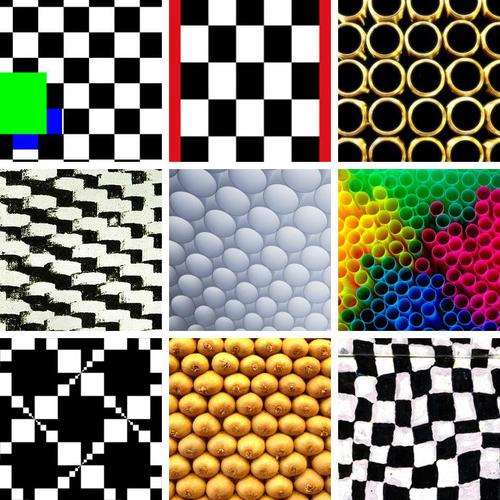}&&
\includegraphics[width=0.23\linewidth]{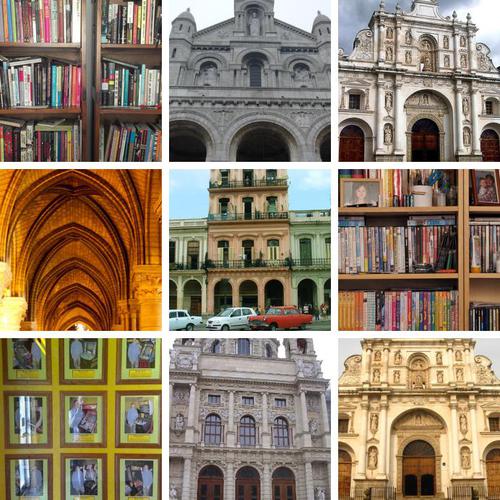}&&
\includegraphics[width=0.23\linewidth]{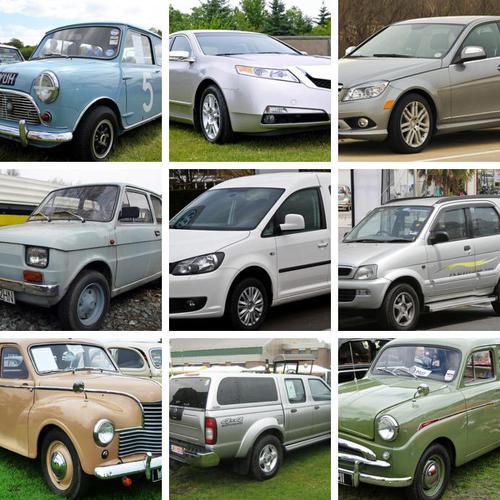}&&
\includegraphics[width=0.23\linewidth]{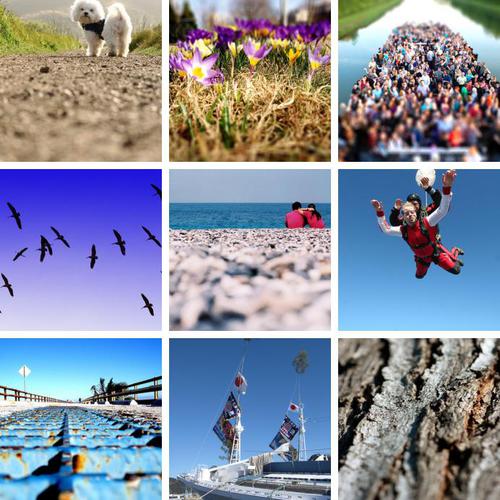}
\end{tabular}
\end{figure}
\begin{figure}[t]
\centering
\begin{tabular}{ccccccc}
Filter $36$ && Filter $41$ && Filter $42$ && Filter $43$
\\
\includegraphics[width=0.23\linewidth]{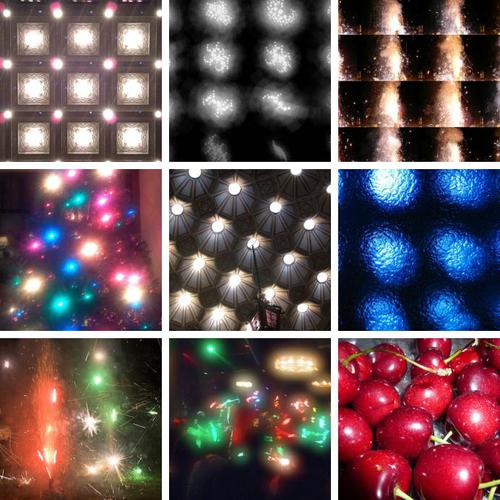}&&
\includegraphics[width=0.23\linewidth]{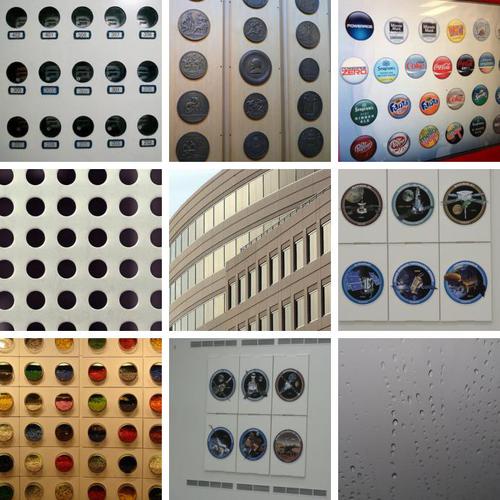}&&
\includegraphics[width=0.23\linewidth]{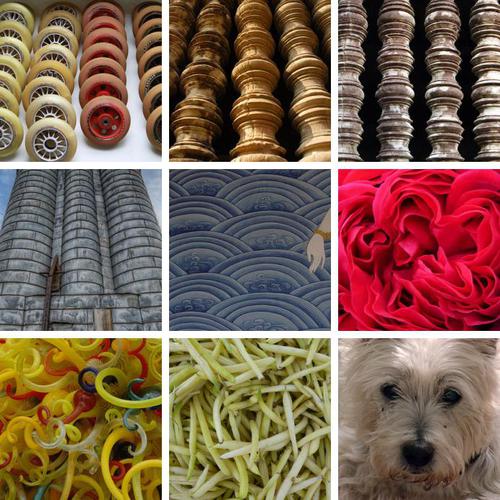}&&
\includegraphics[width=0.23\linewidth]{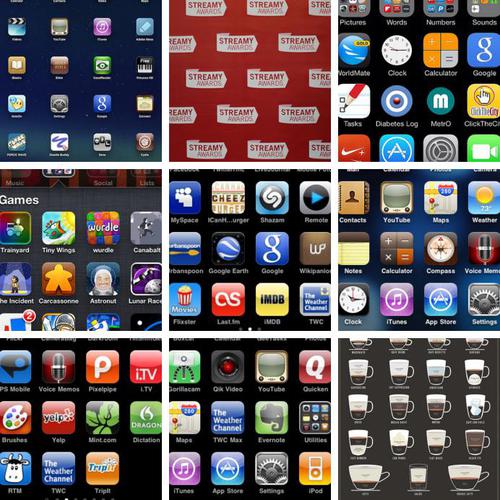}
\\
Filter $45$ && Filter $52$ && Filter $55$ && Filter $56$
\\
\includegraphics[width=0.23\linewidth]{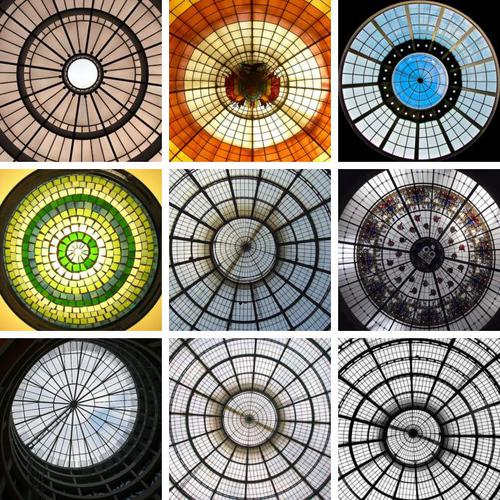}&&
\includegraphics[width=0.23\linewidth]{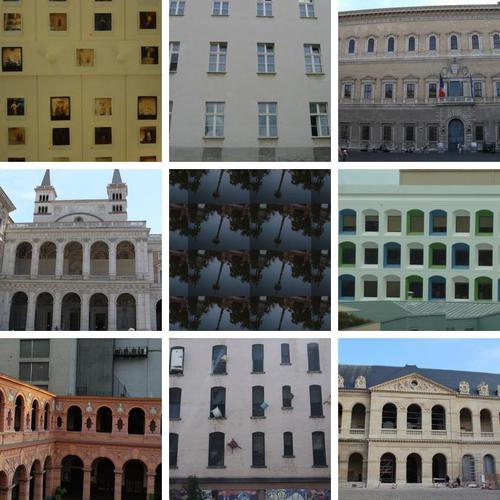}&&
\includegraphics[width=0.23\linewidth]{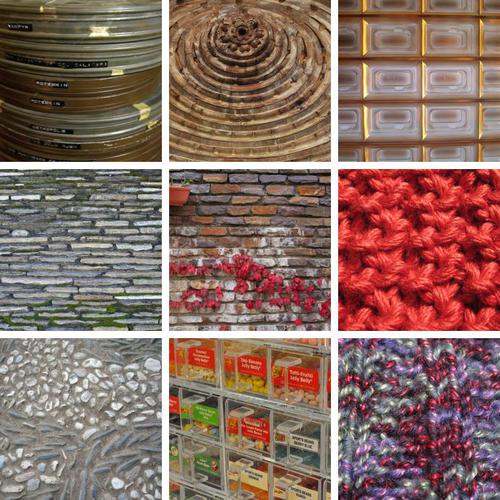}&&
\includegraphics[width=0.23\linewidth]{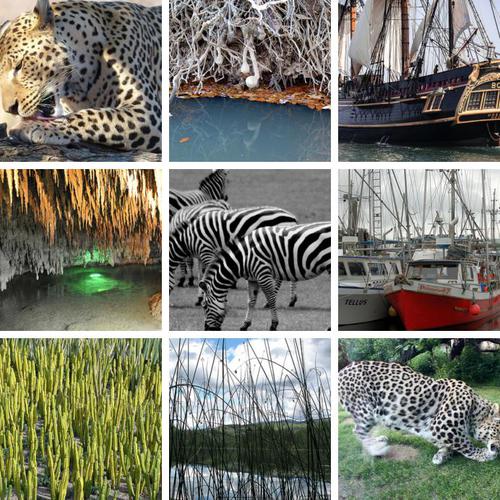}
\\
Filter $57$ && Filter $58$ && Filter $59$ && Filter $60$
\\
\includegraphics[width=0.23\linewidth]{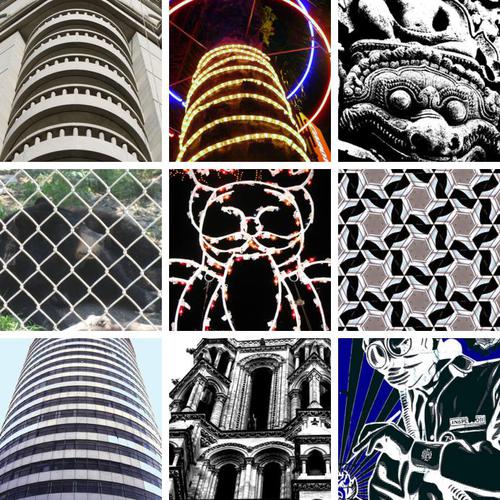}&&
\includegraphics[width=0.23\linewidth]{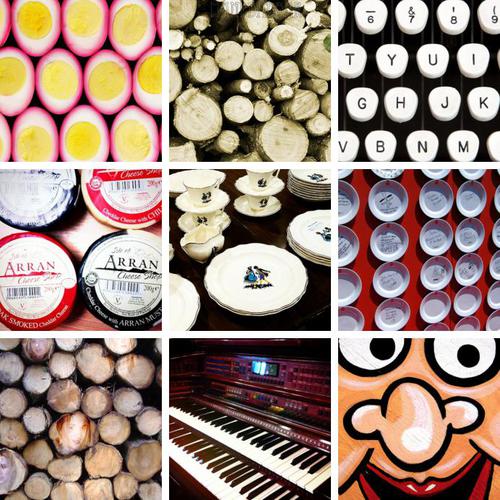}&&
\includegraphics[width=0.23\linewidth]{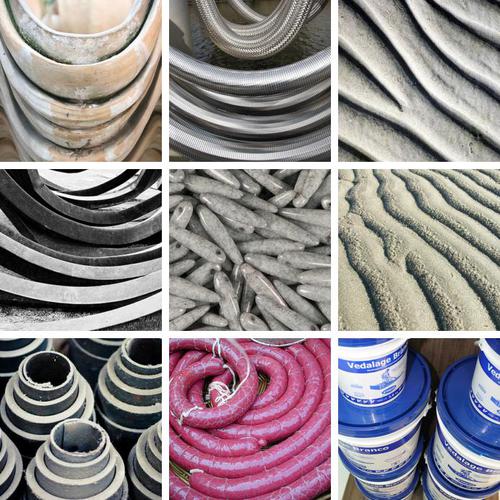}&&
\includegraphics[width=0.23\linewidth]{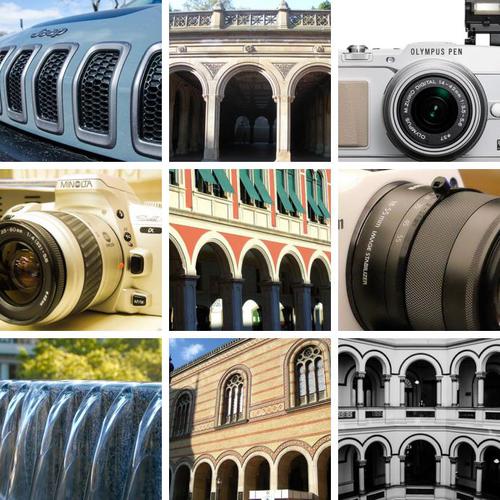}
\\
Filter $62$ && Filter $63$ && Filter $65$ && Filter $66$
\\
\includegraphics[width=0.23\linewidth]{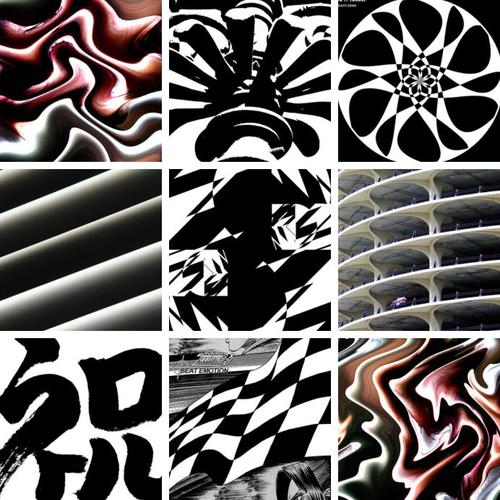}&&
\includegraphics[width=0.23\linewidth]{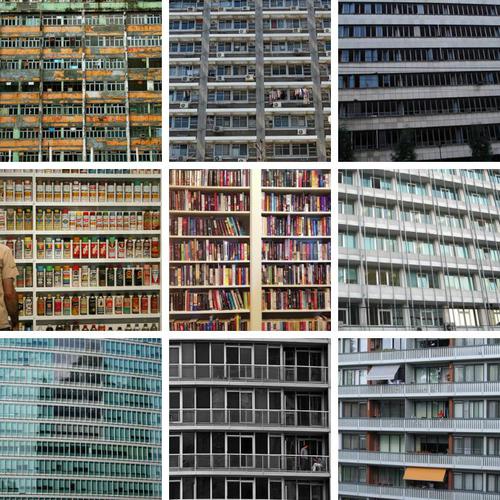}&&
\includegraphics[width=0.23\linewidth]{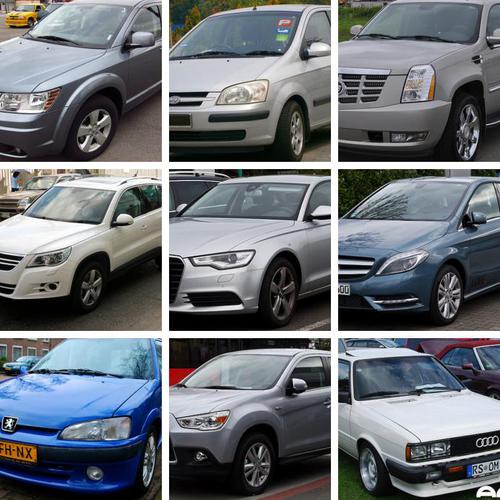}&&
\includegraphics[width=0.23\linewidth]{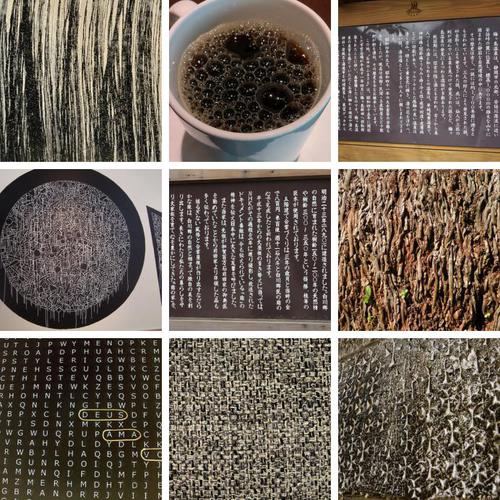}
\\
Filter $67$ && Filter $68$ && Filter $69$ && Filter $70$
\\
\includegraphics[width=0.23\linewidth]{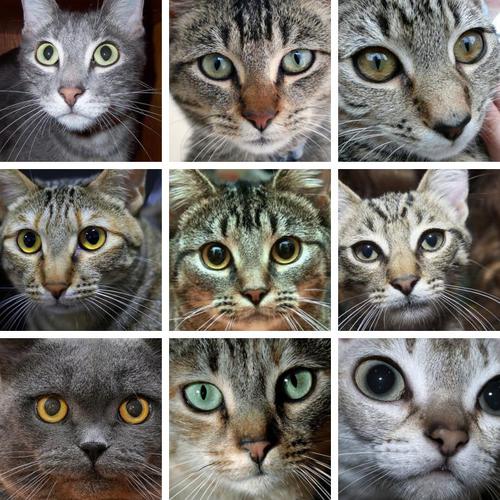}&&
\includegraphics[width=0.23\linewidth]{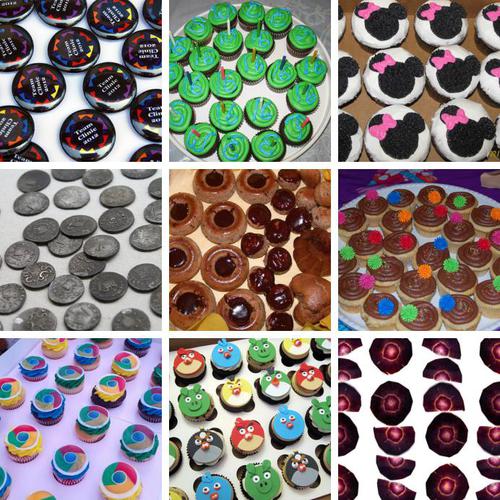}&&
\includegraphics[width=0.23\linewidth]{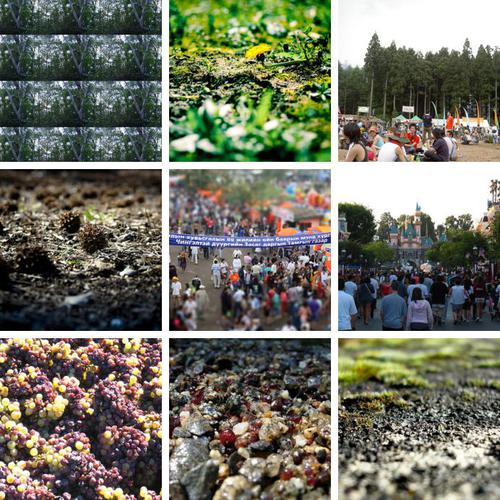}&&
\includegraphics[width=0.23\linewidth]{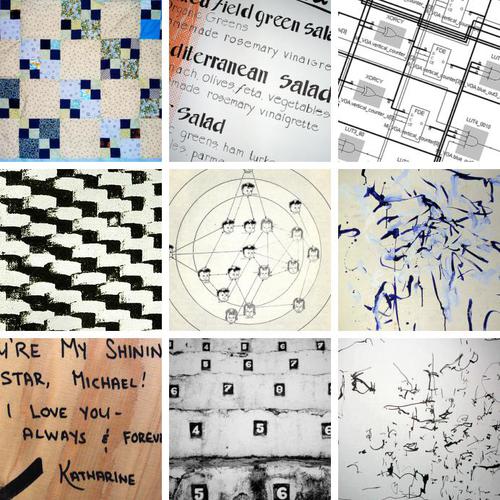}
\\
Filter $74$ && Filter $75$ && Filter $76$ && Filter $78$
\\
\includegraphics[width=0.23\linewidth]{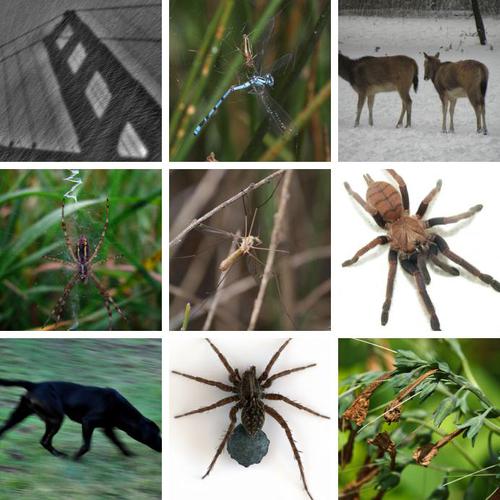}&&
\includegraphics[width=0.23\linewidth]{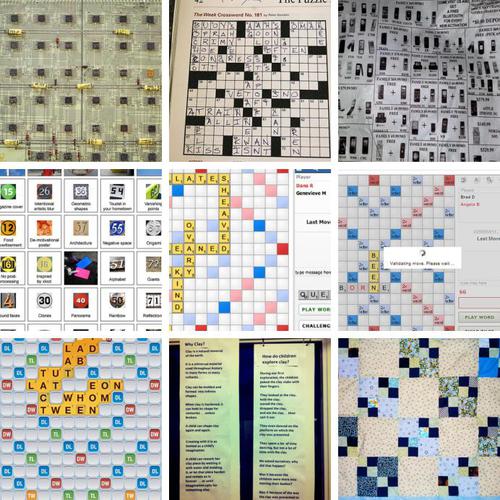}&&
\includegraphics[width=0.23\linewidth]{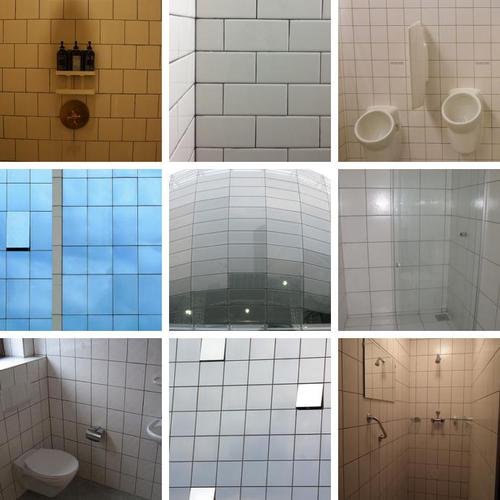}&&
\includegraphics[width=0.23\linewidth]{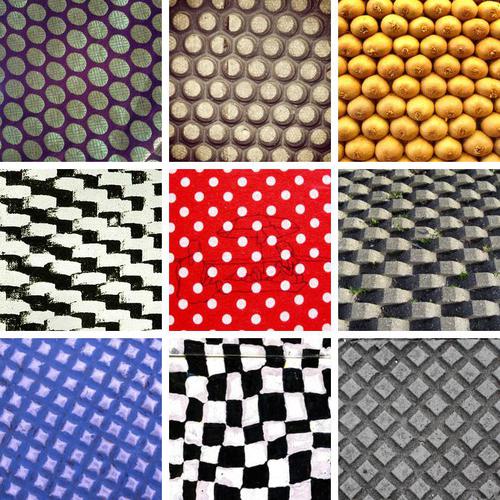}
\end{tabular}
\end{figure}
\begin{figure}[t]
\centering
\begin{tabular}{ccccccc}
Filter $79$ && Filter $81$ && Filter $83$ && Filter $84$
\\
\includegraphics[width=0.23\linewidth]{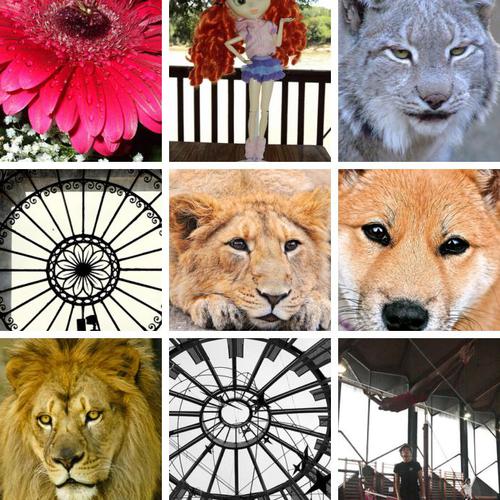}&&
\includegraphics[width=0.23\linewidth]{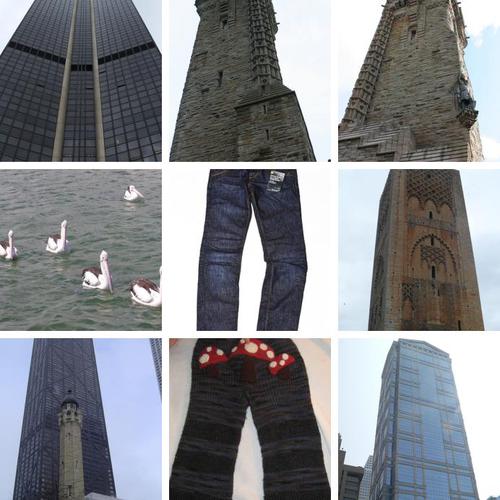}&&
\includegraphics[width=0.23\linewidth]{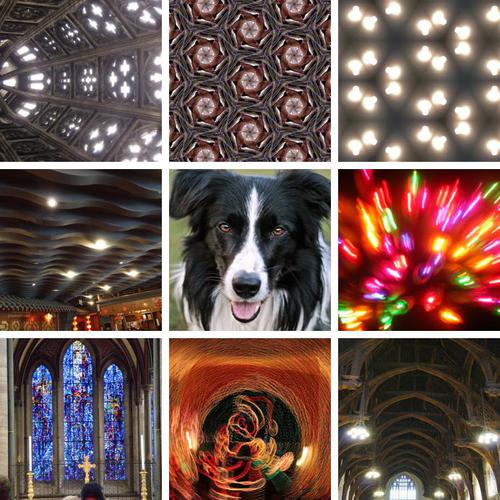}&&
\includegraphics[width=0.23\linewidth]{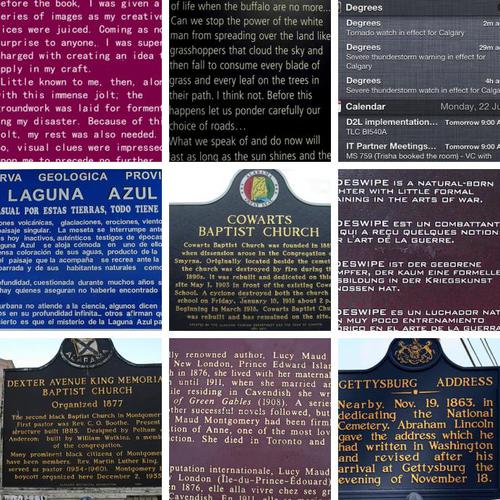}
\\
Filter $85$ && Filter $88$ && Filter $90$ && Filter $91$
\\
\includegraphics[width=0.23\linewidth]{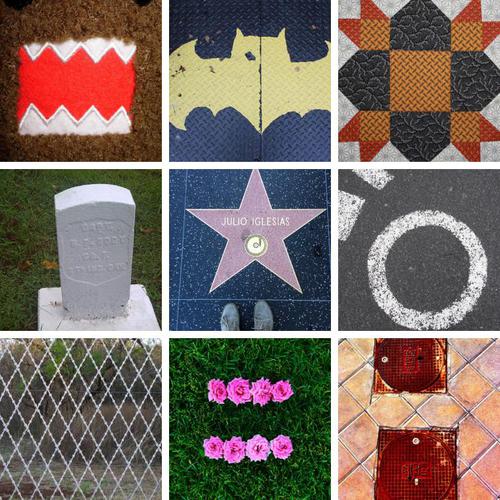}&&
\includegraphics[width=0.23\linewidth]{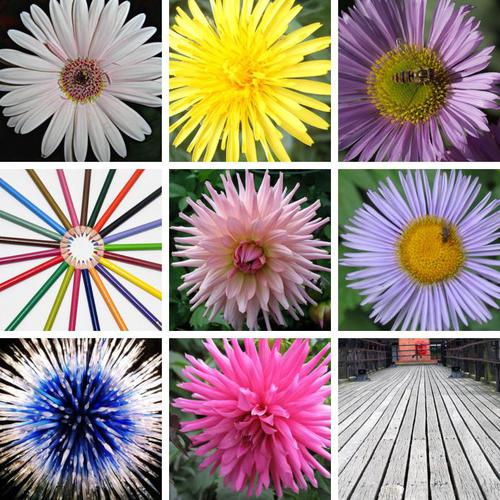}&&
\includegraphics[width=0.23\linewidth]{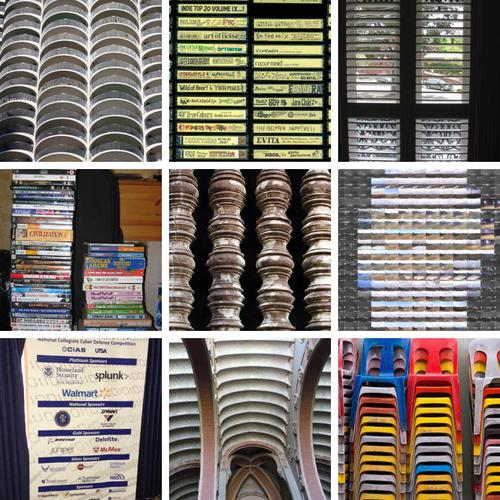}&&
\includegraphics[width=0.23\linewidth]{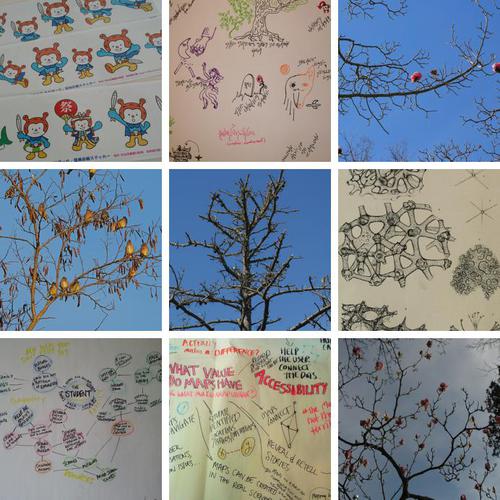}
\\
Filter $92$ && Filter $93$ && Filter $96$ && Filter $97$
\\
\includegraphics[width=0.23\linewidth]{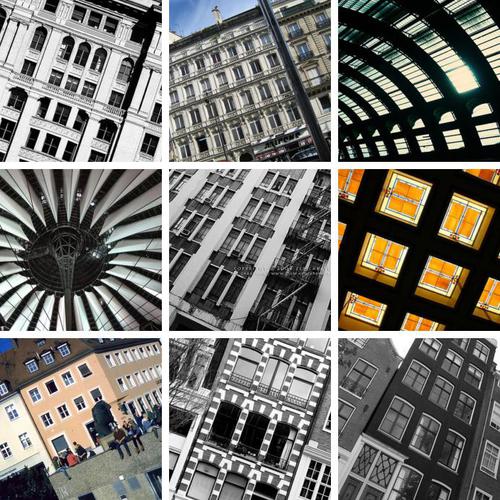}&&
\includegraphics[width=0.23\linewidth]{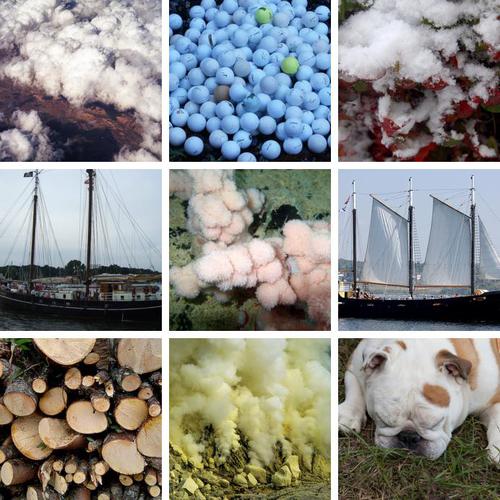}&&
\includegraphics[width=0.23\linewidth]{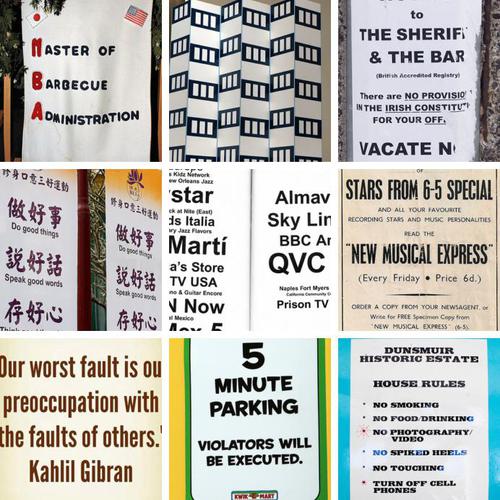}&&
\includegraphics[width=0.23\linewidth]{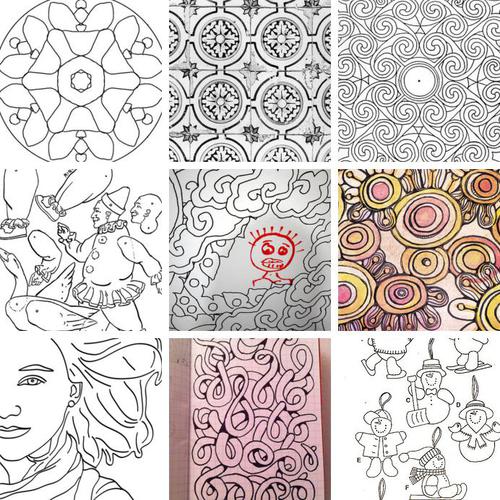}
\end{tabular}
\caption{
  Top $9$ activated images from a random subset of $10$ millions images from YFCC100M for the $100$ first filters in the last convolutional layer of an AlexNet trained with \OURS (we do not display humans).
}
\label{fig:waouh2}
\end{figure}

\section{Erratum [18/03/2019]}

In the original version of the paper, we report classification accuracy averaged over $10$ crops for the linear classifier experiments on Places and ImageNet datasets.
However, other methods report accuracy of the central crop so our comparison wasn't fair.
Nevertheless, it does not change the conclusion of these experiments, our approach still outperforms the state of the art from \texttt{conv3} to \texttt{conv5} layers.
In Table~\ref{tab:erratum}, we show our results both for single and $10$ crops.
\begin{table}[t]
  \centering
  \resizebox{\columnwidth}{!}{%
    \begin{tabular}{@{}l c ccccc c ccccc@{}}
      \toprule
            &~~~& \multicolumn{5}{c}{ImageNet} &~~~& \multicolumn{5}{c}{Places} \\
            \cmidrule{3-7} \cmidrule{9-13}
      Method && \texttt{conv1} & \texttt{conv2} & \texttt{conv3} & \texttt{conv4} & \texttt{conv5} && \texttt{conv1} & \texttt{conv2} & \texttt{conv3} & \texttt{conv4} & \texttt{conv5} \\
      \midrule
      \OURS single crop  && $12.9$ & $29.2$ & $38.2$ & $39.8$ & $36.1$ && $18.6$ & $30.8$ & $37.0$ & $37.5$ & $33.1$ \\
      \OURS $10$ crops  && $13.4$ & $32.3$ & $41.0$ & $39.6$ & $38.2$ && $19.6$ & $33.2$ & $39.2$ & $39.8$ & $34.7$ \\
      \bottomrule
    \end{tabular}
  }
  \vspace{\soustable}
  \caption{
    Linear classification on ImageNet and Places using activations from the convolutional layers of an AlexNet as features.
  }
  \label{tab:erratum}
\end{table}